\documentclass{article} % For LaTeX2e
\usepackage{arxiv_for_iclr,times}

\usepackage{amsmath,amsfonts,bm}

\def\eqref#1{equation~\ref{#1}}
\def\1{\bm{1}}

\DeclareMathAlphabet{\mathsfit}{\encodingdefault}{\sfdefault}{m}{sl}
\SetMathAlphabet{\mathsfit}{bold}{\encodingdefault}{\sfdefault}{bx}{n}

\renewcommand{\eqref}[1]{(\ref{#1})}

\usepackage[colorlinks,breaklinks]{hyperref}
\usepackage{url}
\usepackage{caption} % DO NOT CHANGE THIS AND DO NOT ADD ANY 

\usepackage{algorithm}
\usepackage{algorithmic}
\usepackage{amsmath}
\usepackage{amssymb}
\usepackage{multirow}
\usepackage{graphicx}
\usepackage{booktabs} 
\title{Dynamic Orthogonal Continual Fine-tuning for Mitigating Catastrophic Forgettings}

\author{Zhixin Zhang$^1$\qquad 
Zeming Wei$^{1*}$\qquad
Meng Sun$^1$\thanks{Corresponce to Zeming Wei (\texttt{weizeming@stu.pku.edu.cn}) and Meng Sun (\texttt{sunm@pku.edu.cn}).}
\vspace{5pt}\\
${}^1$Peking University
}

\iclrfinalcopy

\begin{document}

\maketitle

\begin{abstract}
Catastrophic forgetting remains a critical challenge in continual learning for large language models (LLMs), where models struggle to retain performance on historical tasks when fine-tuning on new sequential data without access to past datasets. In this paper, we first reveal that the drift of functional directions during the fine-tuning process is a key reason why existing regularization-based methods fail in long-term LLM continual learning. To address this, we propose Dynamic Orthogonal Continual (DOC) fine-tuning, a novel approach that tracks the drift of these functional directions and dynamically updates them during the fine-tuning process. Furthermore, by adjusting the gradients of new task parameters to be orthogonal to the tracked historical function directions, our method mitigates interference between new and old tasks. Extensive experiments on various LLM continual learning benchmarks demonstrate that this approach outperforms prior methods, effectively reducing catastrophic forgetting and providing a robust tool for continuous LLM fine-tuning. Our code is available at \url{https://github.com/meloxxxxxx/DOC}.
\end{abstract}

\section{Introduction}

Recently, Large Language Models (LLMs) have achieved significant milestones in various tasks based on their extensive capacity and knowledge. In particular, fine-tuning LLMs with task-specific data has emerged as a popular learning paradigm in their diverse applications. In this context, LLM Continual Learning~\citep{llmsurvey}, which fine-tunes LLMs with evolving tasks and data, has become a crucial technique for updating their knowledge to keep pace with new environments and goals. However, a critical challenge of continual learning is catastrophic forgetting~\citep{CL-challenge}, where the model forgets the knowledge it acquired from previous tasks after receiving new updates.

Existing continual learning approaches for LLMs can be categorized into the following types~\citep{llmsurvey, lifelongllmsurvey}: Rehearsal-based~\citep{rehearsal1, rehearsal2, rehearsal3}, Architecture-based~\citep {expert1, expertavg, llama-pro}, Prompt-based~\citep{l2p,lfpt5,progressive-prompt}, and Regularization-based approaches~\citep{ogd, olora, lwf, ewc, synapsis}. While the first three approaches may suffer from significant computational or memory overhead issues (\textit{e.g.}, training additional modules or storing historical data), regularization-based continual learning for LLMs does not suffer from these issues and has been acknowledged as an efficient approach~\citep{llmsurvey}. More formally, we denote that all existing regularization-based methods abide by the following outline: 
\begin{itemize}
    \item[\textbf{Step (1)}.] Record the functional directions, mainly including the \textit{gradient direction} of the model parameter, on historical tasks~\citep{direction1, direction2, direction3};
    \item[\textbf{Step (2)}.] Regularize new updates based on these historical functional directions.
\end{itemize}

For instance, Elastic Weight Consolidation (EWC) methods~\citep{ewc, synapsis} and orthogonal optimization methods, including Orthogonal Gradient Descent (OGD)~\citep{ogd} and Orthogonal Subspace Learning (O-LoRA)~\citep{olora}, employ historical gradient directions and vectors in LoRA matrices of the model for regularization.

\begin{figure}
    \centering

\begin{tabular}{c c}
    %\begin{figure}%[t]
    \centering
   % \begin{tabular}{c}
    \includegraphics[width=0.4\textwidth]{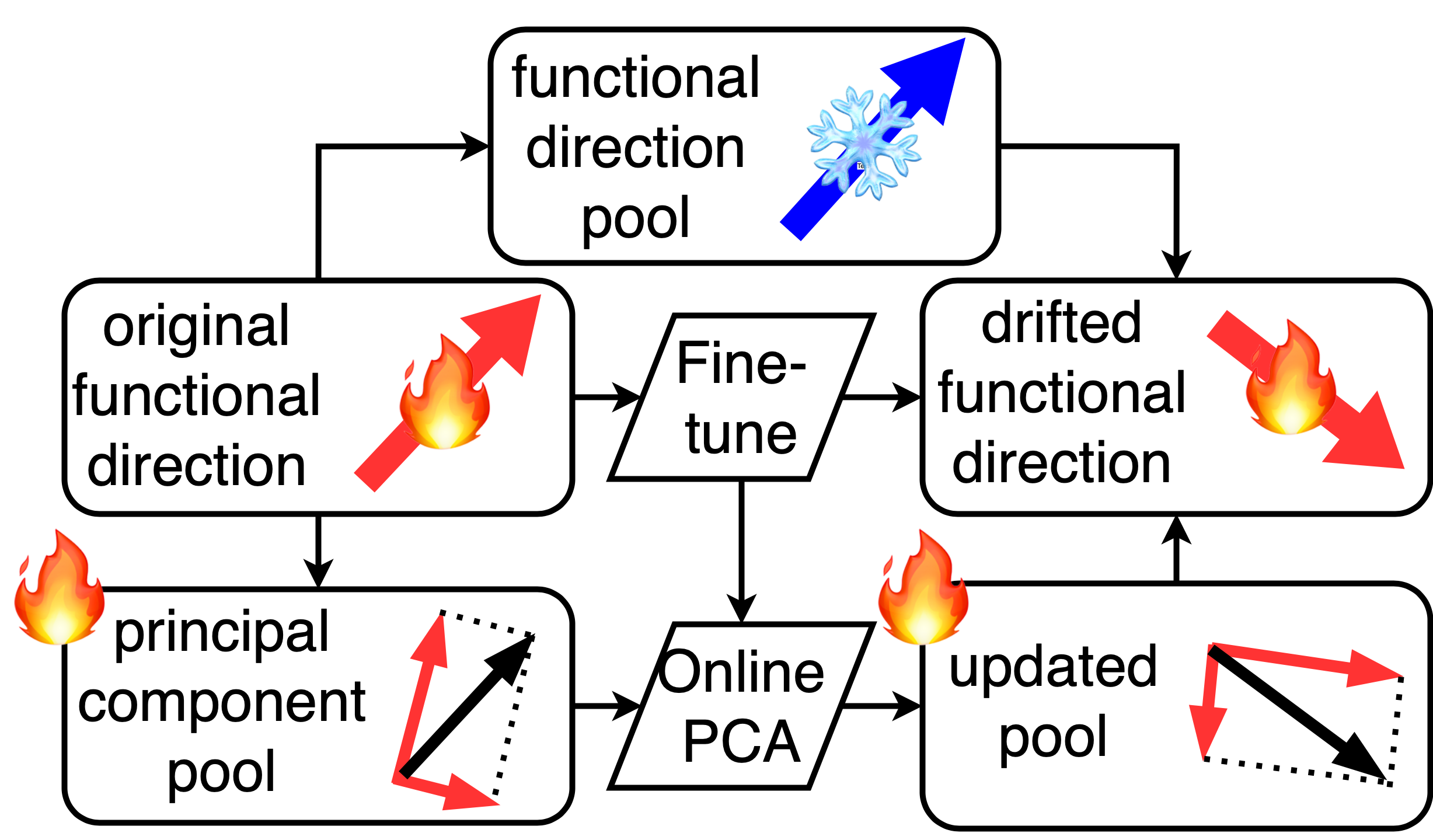} &
    \includegraphics[width=0.53\textwidth]{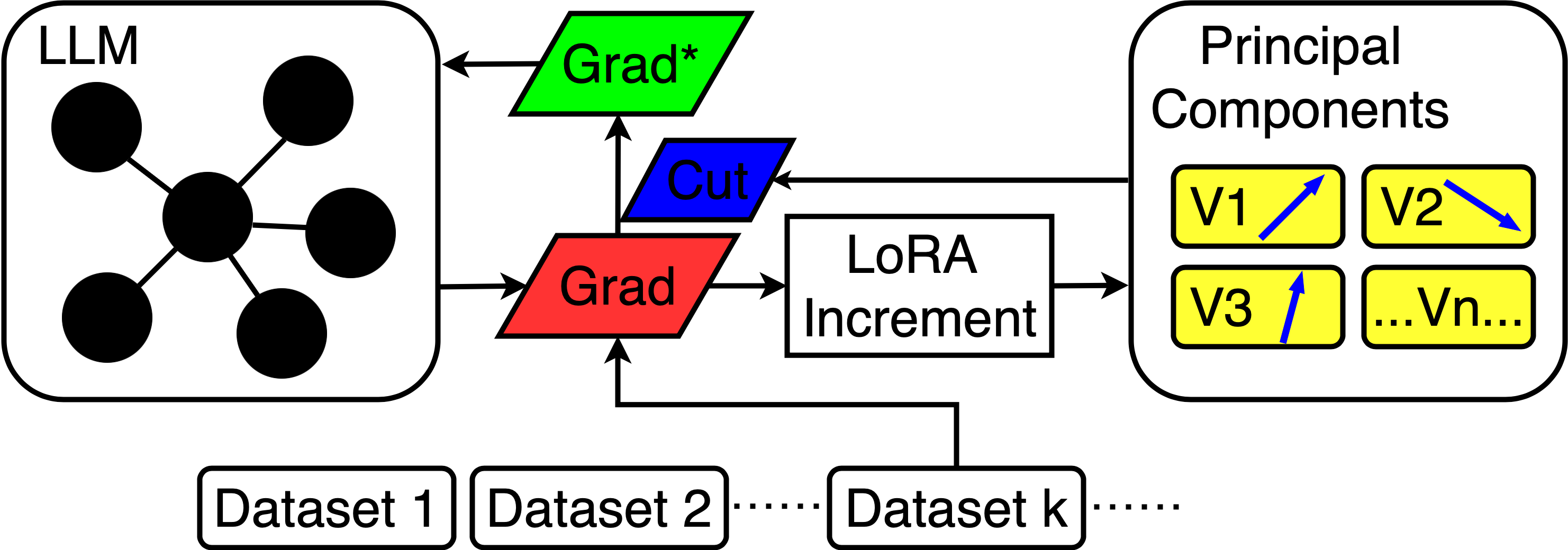} \\

    (a) Drifting functional directions&

    (b) Method overview \\

\end{tabular}
\caption{An introduction to our work.  Prior methods record the functional directions in a \textit{fixed} pool and try to regularize future updates with it, which is shown on the upper half of Figure (a). Our method (the lower half) \textit{updates} these directions with Online PCA for better regularization.
Figure (b) presents an overview of our method. In a sequence of incoming datasets, we compute gradients and LoRA increments to update a set of principal components that represent drifting functional directions. We cut them off from current gradients to avoid forgetting historical functions.}
\label{overview}
\vspace{-10pt}
\end{figure}

However, current regularization-based continual learning still faces the catastrophic forgetting problem, leaving a gap for its practical deployment. In this paper, we aim to mitigate this problem by identifying a key problem in the regularizers. Specifically, we find that the drift of functional directions~\citep{polytope} during continuous fine-tuning poses a significant issue for their regularizations. While functional directions may be valid within a local neighborhood around a static point in the parameter space, continuous fine-tuning can break this locality when moving the model weights towards other spaces, thus destroying the functionality of these directions, as shown in Figure~\ref{overview}(a). This observation is detailed in Section~\ref{motivation}. In the settings of regularization-based methods, the difficulty lies in the lack of access to historical data, which makes it challenging to update their functional directions in the current parameter space.

Based on the observation above, we propose our method that tracks the drifting functional directions of historical tasks with the latest task data. Since LLMs primarily fine-tune within a low-rank subspace~\citep{intrinsiclora}, all tasks share most of the functional directions in this subspace with different linear combinations. Thus, we employ a modified Online Principal Component Analysis~\citep{pcasurvey} to extract these directions from their combinations to capture and track the evolving functional directions. Leveraging these up-to-date functional directions, we cut gradients of new task parameters to be orthogonal to the tracked historical function directions, following prior orthogonal methods including OGD~\citep{ogd} and O-LoRA~\citep{olora}, which mitigates the interference between new and old tasks. However, a key difference between our method and other orthogonal methods is that we dynamically \textit{update} the functional directions rather than regularizing on \textit{fixed} ones. Tracking these functional directions, which prior works often overlook, is crucial for preserving functions that lie in drifting directions. A brief overview of our method is in Figure~\ref{overview}(b).

Extensive experiments verify the drift of functional directions and demonstrate the effectiveness of our method in tracking them, offering a substantiated motivation for our method. Furthermore, experiments on various LLM continual learning benchmarks demonstrate that our approach significantly mitigates the catastrophic forgetting issues in online streaming data scenarios, and outperforms prior methods, \textit{e.g.}, we respectively achieve an accuracy of 77.7 and 73.4 in standard CL benchmark~\citep{clbenchmark} and long chains of tasks for LLaMA-7B~\citep{llama}, compared to 76.5 and 71.9 of O-LoRA~\citep{olora}, the previous state-of-the-art regularization-based method. In summary, our contributions are as follows:
\begin{itemize}
\item We reveal the drift of function directions in the fine-tuning process, which explains why regularization-based approaches fail in long-term LLM continual learning.
\item Based on this discovery, we propose the Dynamic Orthogonal Continual Fine-tuning (DOC) method that tracks the drift of functional directions to mitigate catastrophic forgetting issues.
\item We conduct extensive experiments to validate that DOC outperforms prior methods in various LLM continual learning benchmarks, contributing an effective tool in continuous LLM fine-tuning.
\end{itemize}

\section{Preliminaries}
\subsection{Continual Learning Setup}
Continual learning for LLMs~\citep{llmsurvey,lifelongllmsurvey} is crucial for updating their knowledge and keeping pace with new goals. In a continual learning scenario, a pre-trained LLM is fine-tuned on an online stream of tasks with their task-specific data. Due to factors like storage costs and privacy protection, historical data cannot be accessed when fine-tuning on the latest one.  

\noindent\textbf{Definition of continual learning}.
Given a LLM $F_\theta$ with parameters $\theta$, a sequence of labeled datasets $\{D_1, D_2, ..., D_N\}$, where $D_t=\{(x_t^i, y_t^i)\}_{i=1}^{n_t}\quad (t = 1,\dots , N)$. Then $F_\theta$ is sequentially fine-tuned on ${D_1, D_2, ..., D_N}$. When fine-tuning on $D_T$, historical datasets, \textit{i.e.} $\{D_1, D_2,...,D_{t-1}\}$, cannot be accessed. The target is an $F_\theta$ that behaves well on all datasets: 
\begin{equation}\label{cltarget}
    \arg\min_{\theta}\sum_{t=1}^N \sum_{i=1}^{n_i} L_t(F_\theta(x_t^i), y_t^i),
\end{equation}
where $L_t$ is the task-specific loss function of the $t$-th task. \textbf{Note that for concision, we substitute $L$ for $L_t$ when fine-tuning on $D_t$ in the following statements}.

\subsection{Low-Rank Adaptation (LoRA)}
When fine-tuning LLMs for specific tasks, there exists a low intrinsic dimension for the parameter update of the model~\citep{intrinsiclora}.
For a weight matrix $W_{m \times n}$ of a pre-trained LLM, LoRA~\citep{lora} employs low-rank matrixes $B_{m \times r}$ and $A_{r\times n}$ ($r \ll min(m,n)$) 
to constrain its update by representing it with a low-rank decomposition:
\begin{equation}\label{lora}
    W^* = W + BA,
\end{equation}
where $W^*$ is the new parameter after fine-tuning. As a result, the propagation process is modified:
\begin{equation}\label{loraincrement}
    W^*x = (W+BA)x = Wx+BAx,
\end{equation}
where $x$ is the input to the module with parameter $W$.

\section{Motivation and the Proposed Method}\label{methods}
In this section, we first reveal that the drift of functional directions is the key issue for existing regularization-based methods in~\ref{motivation}, then propose a method to track drifting functional directions and validate the effectiveness of our tracking method in~\ref{tracking method}, and finally cut the parameter increment of new tasks to be orthogonal to historical ones in~\ref{cutting method}. 

\subsection{Motivation: Analysis of Existing Regularization Methods}\label{motivation}
Our method is developed using a regularization-based approach in consideration of its little computational or memory overhead issues~\citep{llmsurvey}. 
While prior research~\citep {lifelongllmsurvey} has demonstrated that existing regularization methods are efficient on short task sequences, their performance is relatively limited in long sequences, leaving a gap for their practical deployment.
In the following parts, we propose an analysis to identify the primary cause of this defect. 

\noindent\textbf{Intrinsic functional directions of LLMs}.
Functional directions~\citep{direction1, direction2, direction3} have become prevalent in research on LLMs. In this paper, we define functional directions of LLMs as \textit{the gradient direction of model parameters on certain datapoints}. Most of the prior regularization-based approaches employ functional directions to approximate the functional unit of certain tasks in LLMs. They adhere to the outline for recording functional directions and regularizing new updates on historical directions.
Specifically, Orthogonal methods, including Orthogonal Gradient Descent (OGD)~\citep{ogd} and Orthogonal Subspace Learning (O-LoRA)~\citep{olora}, avoid perturbing the historical settings of the model through orthogonal approaches, and the two respectively employ gradient directions and LoRA vectors as the regularized functional directions.
Elastic Weight Consolidation(EWC) methods~\citep{ewc, synapsis} employ the Fisher information matrix for its consolidation, which is also computed with gradients.

\noindent\textbf{Functional directions drift in the fine-tuning process}.
In this part, we identify that the drift of functional directions during the continuous fine-tuning process is the key issue of the regularizations above.
Specifically, in the process of continually fine-tuning an LLM, the locality of linearity in its deep neural networks is broken~\citep{polytope}, thus destroying the functionality of the directions extracted in earlier steps. 
Consequently, regularization in these directions deviates from the original purpose in the continual fine-tuning, as demonstrated in  Figure~\ref{overview}(a). In this part, we present the following observations regarding the drifts proposed above. We take fine-tuning Llama-2~\citep{llama} on CL Benchmark~\citep{clbenchmark} as the example in this experiment, and measure the drift of the gradient direction during continual fine-tuning.

\begin{figure}[t]
    \centering
    \begin{tabular}{c c}
    \includegraphics[width=0.45\textwidth]{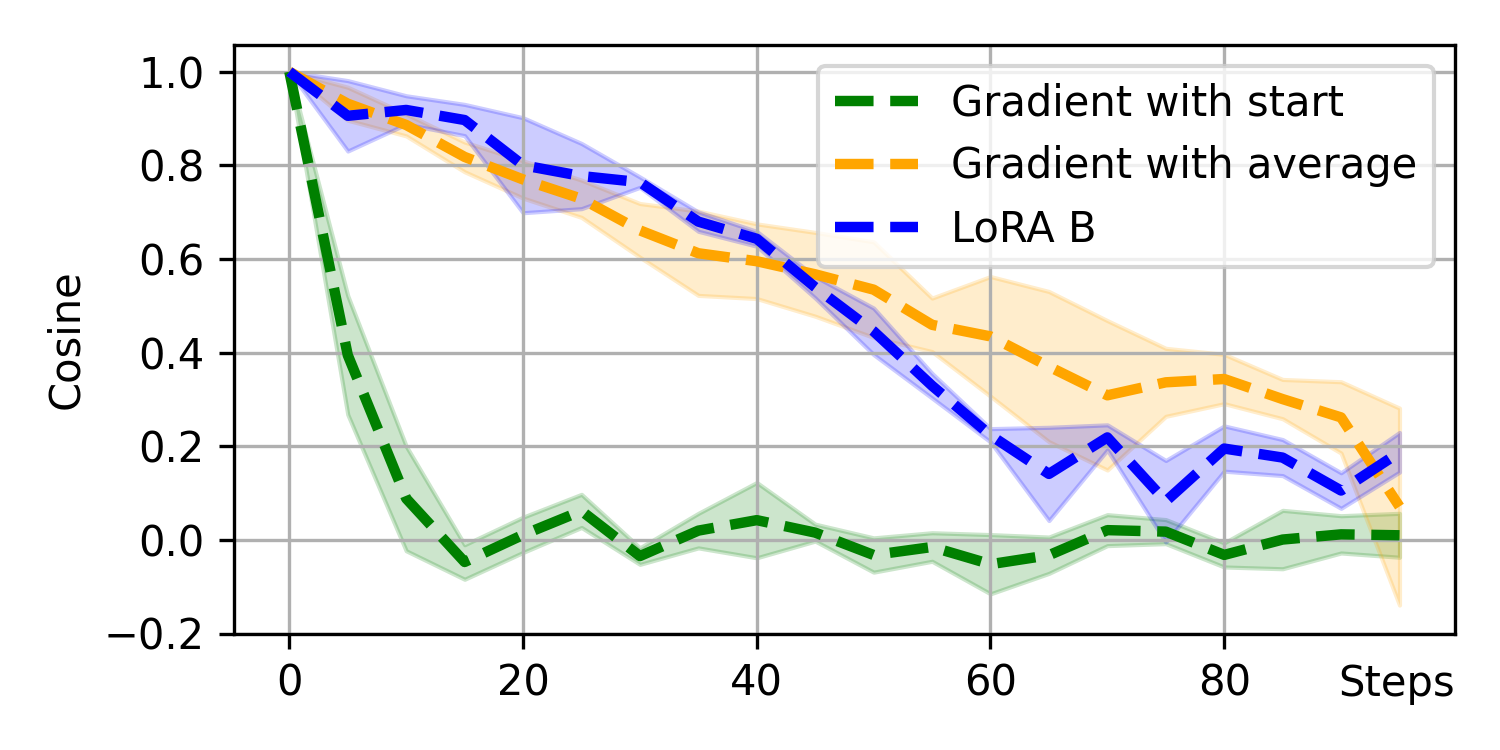} & \includegraphics[width=0.45\textwidth]{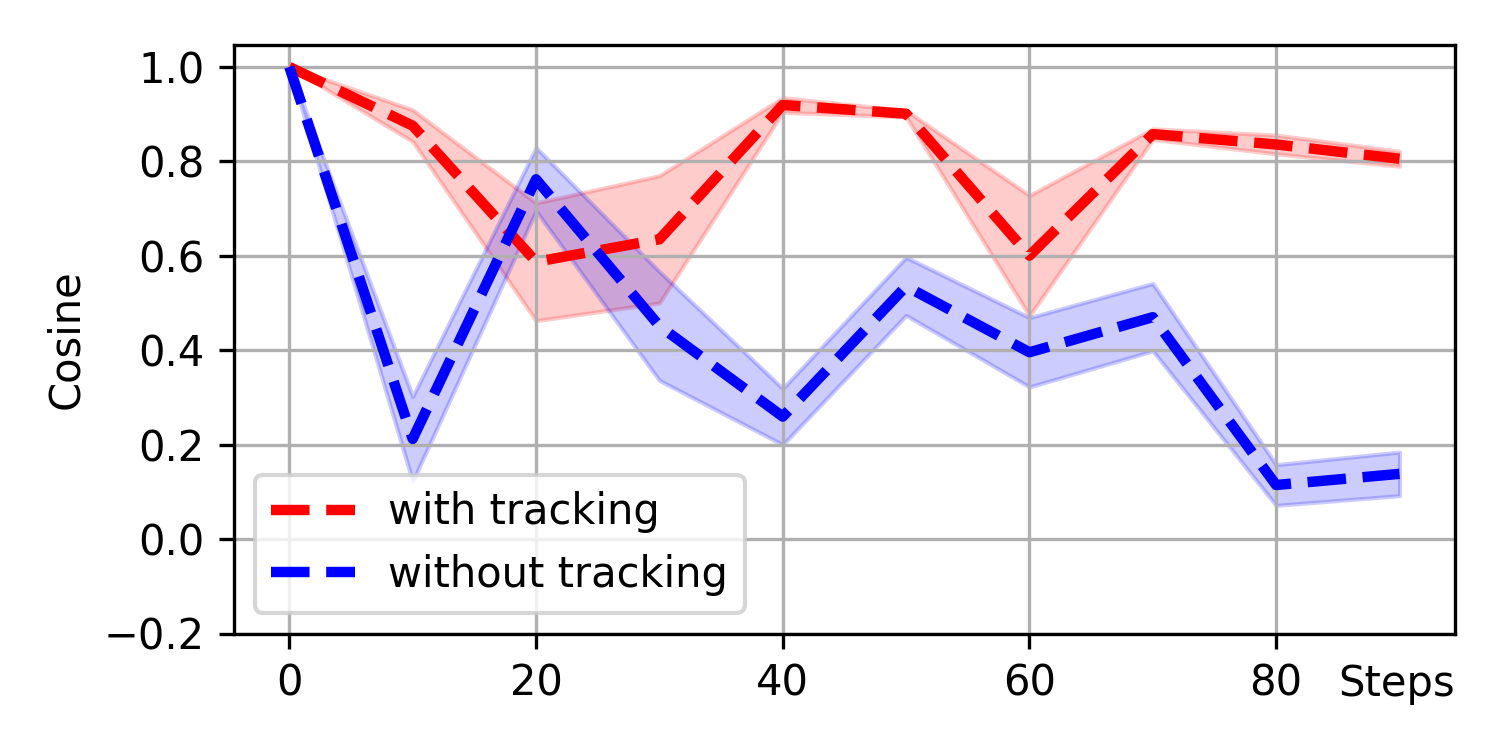} \\
    (a)  & (b) 
    \end{tabular}
    %\vspace{-10pt}
    \caption{Quantification of functional direction drift  regarding a particular datapoint $(x,y)$. Figure~(a) shows the cosine similarity between current and historical functional directions. The green line shows 
    $\textbf{cos}\langle G_T, G_1 \rangle$, where $G_T = \nabla_{\theta} L(F_{\theta_T}(x), y)$, $\theta_T$ is the model parameter in the $T$~th fine-tuning step. The yellow line shows 
    %the similarity of the gradient with its average, \textit{i.e.} 
    $\textbf{cos} \langle G_T, \bar{G}_T \rangle$, where $\bar{G}_T = \frac{1}{T}\sum_{t=1}^TG_t$. The blue line shows the average similarity of  $\beta_1, \beta_2,...,\beta_r$ in LoRA $B$ matrices with their start value, \textit{i.e.} $\frac{1}{r}\sum_{n=1}^r \textbf{cos} \langle \tilde{\beta}_n, \beta_n \rangle$, where $\tilde{\beta}$ is the current one, $\beta$ is the start one. 
    Figure~(b) shows the effect of tracking functional directions. We initialize the principal components during the first dataset, and measure the drift in the following steps.
    The red line shows the drift with $ \textbf{cos}\langle\textbf{coord}(h^*_T), \textbf{coord}(h^*_1)\rangle$, where $h^*_T$ is the LoRA increment~(computed with Equation~\eqref{onlinecl-data})
    in the $T$-th step.
    For contradiction, we freeze the update of principal components (the blue line). The results are the average of randomly-chosen datapoints, with standard deviation shown.}
    %\vspace{-10pt}
    \label{drift-quantification}
\end{figure}

As shown in Figure~\ref{drift-quantification}(a), with the fine-tuning process conducted, the functional directions captured earlier no longer represent the current ones, exposing the ineffectiveness of employing \textit{fixed} singular or average gradient as the functional direction, which is conducted in EWC~\citep{ewc, synapsis} and OGD~\citep{ogd}. %这里点名好像不太好
Similarly, we also investigate the drift of column vectors in LoRA $B$ matrices employed by O-LoRA~\citep{olora}, \textit{i.e.} $\beta_1, \beta_2,...,\beta_r$ in $B=(\beta_1, \beta_2,...,\beta_r)$. %Please note that the notation of the O-LoRA paper is $W^*=W+AB$ for LoRA Equation~\eqref{lora}, which is different from ours. 
The results (blue line, denoted as LoRA $B$) show that employing column vectors in LoRA $B$ matrices mitigates the loss of functional directions. However, it does not resolve the drift issue fundamentally. 
Overall, we identify that the prevalent problem in prior regularization-based methods is the drift of functional directions,  
%The efficacy of regularizing on fixed directions deviates from the purpose of preserving its original functionalities as they drift away, 
indicating that we need to dynamically update the functional directions rather than relying on fixed ones.

% Overall, we can 
% 这里需要一个简单的总结

\subsection{Tracking the drift of functional directions}\label{tracking method}

The difficulty of mitigating the drift of functional directions lies in updating the
functional directions of historical data in the current parameter space, as there is no access to historical data in the settings of regularization-based methods. 
To tackle this issue, we propose our method to track the drifting functional directions of historical tasks with the latest task data.

As shown in Equation~\eqref{loraincrement}, LLMs primarily fine-tune within a low-rank subspace, \textit{i.e.} the space of $BAx$, such that all tasks share most of the bases in this subspace, and the functional directions are different linear combinations of these bases. By extracting and updating the bases from the functional directions of the current task, we update the shared bases of historical functional directions, thus updating historical functional directions themselves. 

\noindent\textbf{Tracking method overview}. 
To achieve the conception above, we select the LoRA increment as the functional direction, and employ Principal Component Analysis (PCA) to extract the bases. The following parts propose respective elaborations.

\noindent\textbf{LoRA increment as functional directions}.
Following prior regularization-based methods, including OGD\citep{ogd} and O-LoRA\citep{olora}, we extract fine-tuning increments as the functional directions of certain continual learning tasks. 
More specifically, the functional direction we select is the increment of LoRA in Equation~\eqref{loraincrement}, that is:
\begin{equation}
\label{onlinecl-gradientdata}
    \mathbf{d}(W_mx_m) = \mathbf{d}(B_mA_mx_m)=\mathbf{d}(B_mA_m)x_m
    =(\mathbf{d}B_m)A_mx_m+B_m (\mathbf{d}A_m)x_m \triangleq p_m,
\end{equation}
where $x_m$ is the input \textit{vector} 
to the $m$-th LoRA module with parameter $B_m$ and $A_m$. Let $\alpha$ be the learning rate, then $\mathbf{d}B = \alpha \nabla_BL$, $\mathbf{d}A = \alpha \nabla_AL$, $L$ is the task-specific loss function. 
We represent the update direction of LoRA with the following concatenated vector:
\begin{equation}\label{onlinecl-data}
    h = \textbf{concat}(p_1,p_2,...,p_M),
\end{equation}
where $M$ is the number of LoRA modules. The concatenation captures the relation between the LoRA increment of different layers. More computational details on $x$ and $h$ are shown in Appendix~\ref{appendix computational details}.

\noindent\textbf{Online PCA}.
To extract the basis of functional directions from their linear combinations, we employ the Online Principal Component Analysis (Online PCA)~\citep{pcasurvey}, which requires only the latest data in memory, conforming to the settings of regularization-based continual learning. 

The target of Online PCA is as follows. Let $\{h_1, h_2,...,h_n\}$ be  functional directions computed with Equation~\eqref{onlinecl-data} on a sequence of incoming data. On receiving a new functional direction $h_t$, Online PCA seeks to update principal components $\{v^1_t, v^2_t,..., v^{K_t}_t\}$ as the basis of functional directions $\{h_1, h_2,...,h_t\}$.
Moreover, when processing the latest data $h_t$, there is no access to historical datas $\{h_1, h_2,...,h_{t-1}\}$. 
This realizes our goal of updating historical functional directions with current ones, $\textit{i.e.}$ updating the representation of historical functional directions with the current functional direction.
Please refer to Algorithm~\ref{method-algorithm} for a summary and  Figure~\ref{overview}(a) for a brief demonstration.

There are multiple approaches to implement Online PCA, including 
Incremental PCA~\citep{incrementalpca1, incrementalpca2} and stochastic approximation methods~\citep{stochasitcpca1, stochasticpca2, stochasticpca3, stochasticpca4}. 
Our method draws inspiration from Candid Covariance-free Incremental PCA (CCIPCA)~\citep{ccipca}, since its edge lies in the ability to add components freely, which is suited for emerging new tasks. It also has a lower computational overhead compared to other techniques. Please refer to Appendix for more technical details on our Online PCA method.

\noindent\textbf{The effectiveness of tracking}.
To evaluate the effectiveness of tracking, we investigate the drift of the functional direction in the subspace of the updated principal components. Specifically, we compute the LoRA increment $h^*_T$ of a particular datapoint in the $T$-th fine-tuning step, and compute its coordinate 
in the subspace of extracted principal components, that is 
\begin{equation}\label{validation-exp-coord}
    \textbf{coord}(h^*_T) = \left((h^*_T, v^1_T), (h^*_T, v^2_T),...,(h^*_T, v^K_T)\right).
\end{equation}
where $(h^*_T,v^k_T) = \frac{h^*_T \cdot v^k_T}{\Vert v^k_T \Vert}$ is the projection of $h^*_T$ on $v^k_T$. 
As shown in Figure~\ref{drift-quantification}(b), by tracking principal components, drifting functional directions can be followed and thus remain in correspondence with their original states; if we forbid tracking, functional directions are gradually lost.

\subsection{Cut fine-tuning directions for function preservation}\label{cutting method}

\begin{algorithm}[tb]
\caption{DOC~(Our method)}
\label{method-algorithm}
\textbf{Input}: Model $F_{\theta}$, where $\theta=(A,B)$ includes LoRA $A,B$ modules; learning rate $\alpha$; the $t$-th incoming dataset $D_t$, expected maximum principal component number $K$ for each new task\\
\textbf{Initialization}: Principal components $v^1_T, v^2_T,...,v^{K_{T}}_T$ extracted from historical fine-tunings, $T$ is the number of finished fine-tuning steps.
\\
\textbf{Output}: Fine-tuned parameter $\theta^*$
\begin{algorithmic}[1] %[1] enables line numbers
\FOR{data point(batch) $(x_t^i,y_t^i)$ in $D_t$}
\STATE extract gradients:  
$\nabla_{B} L = \nabla_{B} L(F_{\theta}(x_t^i),y_t^i) \quad 
    \nabla_{A} L = \nabla_{A} L(F_{\theta}(x_t^i),y_t^i)$
\STATE compute current LoRA increment $h_{T+i}$ with Equation~\eqref{onlinecl-data}
\STATE use $h_{T+i}$ to update principal components with Online PCA Algorithm on the basis of existing $v^1_{T+i-1}, v^2_{T+i-1},...,v^{K_{T+i-1}}_{T+i-1}$, get  $v^1_{T+i}, v^2_{T+i},...,v^{K_{T+i}}_{T+i}$ $(K_{T+i-1} \leq K_{T+i} \leq K\cdot t)$

\STATE cut $\nabla_B L$ with Equation~\eqref{cut}, get $(\nabla_B L)_\text{cut}$

\STATE update parameter:
$
    B = B-\alpha \cdot (\nabla_B L)_\text{cut}  \quad A = A-\alpha \cdot \nabla_A L
$
\ENDFOR

\RETURN $\theta^*  =(A,B)$

\end{algorithmic}
\end{algorithm}

Following prior orthogonal space fine-tuning approaches, including OGD~\citep{ogd} and O-LoRA~\citep{olora}, for regularization-based continual learning, we try to make the parameter increment of new tasks orthogonal to historical ones. This avoids changing the functional directions representing historical tasks, thus protecting historical functions. 
Specifically, we make the current LoRA increment $h_T$ orthogonal to historical ones, whose basis are principal components $\{v^1_T, v^2_T, ..., v^K_T\}$. 
The goal is as follows:
\begin{equation}\label{cut-goal}
    h_T \perp v_T^k \quad k=1,2,...,K.
\end{equation}
Note that in Equation~\eqref{onlinecl-data} we have
\begin{equation}
    h_T = \textbf{concat}\left(\mathbf{d}(B_mA_mx_m) \quad m=1,2,...,M \right),
\end{equation}
so we disassemble the concatenation to realize the orthogonality in Equation~\eqref{cut-goal}. The disassembly is as follows:
\begin{equation}
    v_T^k = \textbf{concat}(v_T^k(m)\quad m=1,2,..., M).
\end{equation}
Then we only need to make
\begin{equation}\label{orth-goal}
    \mathbf{d}(B_mA_mx_m) \perp v_T^k(m)  \quad m=1,2,...M \quad k=1,2,...,K
\end{equation}
\textbf{Note that we substitute $BAx$ for $B_mA_mx_m$ and $\tilde{v}^k$ for $v_T^k(m)$ in the following statements for concision}. As $\mathbf{d}(BAx) = (\mathbf{d}B)Ax+B(\mathbf{d}A)x$, we realize the orthogonality in Equation~\eqref{orth-goal} respectively for $(\mathbf{d}B)Ax$ and $B(\mathbf{d}A)x$:
\begin{equation*}
    (\mathbf{d}B)Ax \perp \tilde{v}^k, \quad B(\mathbf{d}A)x \perp \tilde{v}^k \quad \text{for}\quad k=1,2,...,K.
\end{equation*}
For $(\mathbf{d}B)Ax$, note that
\begin{equation}
    (\mathbf{d}B)Ax = (\mathbf{d}\beta_1,\mathbf{d}\beta_2,..., \mathbf{d}\beta_r)(Ax)
    %\Rightarrow (\mathbf{d}B)Ax 
    \in \langle \mathbf{d}\beta_1,\mathbf{d}\beta_2,...,\mathbf{d}\beta_r \rangle \label{disensemble}
\end{equation}
So we only need to cut $\mathbf{d}\beta_i = \alpha \cdot \nabla_{\beta_i} L \quad (i=1,2,..,r)$ to be orthogonal to $\tilde{v}^k \quad (k=1,2,...,K)$. That is:
\begin{equation}
    \nabla_{\beta_i}L \perp \tilde{v}^k  \quad i=1,2,...,r  \quad k=1,2,...,K
\end{equation}
Then we reach the following gradient cut:
\begin{equation}\label{cut}
    (\nabla_{\beta_i} L)^* = \nabla_{\beta_i} L - \sum_{k=1}^K 
    \frac{\nabla_{\beta_i} L \cdot v_T^k}{\Vert v_T^k \Vert ^2} \cdot v_T^k \quad i=1,2,...,r.
\end{equation}
Now we get $\left(\nabla_{B}L)_\text{cut} = ((\nabla_{\beta_1} L)^*, (\nabla_{\beta_2}L)^*,..., (\nabla_{\beta_r}L)^*\right)$. 
Note that the cut above removes the correlation with input $x$ since Equation~\eqref{disensemble}, making the orthogonality hold true for all kinds of input $x$. This is significant in preserving historical functions on all tasks.

For $B(\mathbf{d}A)x$, assume that we have employed $(\nabla_{B}L)_\text{cut}$ in previous steps, then their aggregated $B = (\beta_1, \beta_2,..., \beta_r)$ satisfies the orthogonality for the former steps. Similar to Equation~\eqref{disensemble}, we have 
\begin{equation}
    B(\mathbf{d}A)x \in \langle \beta_1, \beta_2,..., \beta_r \rangle ,
\end{equation}
so the orthogonality holds for $B(\mathbf{d}A)x$. We keep $B(\mathbf{d}A)x$ intact as a momentum for optimization, which means we keep the original $\mathbf{d}A$ and $\nabla_AL$.

Please note that the above orthogonal cut does not harm the gradient descent, as described in the paper of OGD~\citep{ogd}. In summary, our complete method is formulated as Algorithm~\ref{method-algorithm}. Please refer to Figure~\ref{overview}(b) for a brief demonstration.

\begin{table*}[!ht]
    \centering
     %\vspace{-5pt}
    \caption{Average Accuracy (AA) of different continual methods on LLaMA-7B.}
    %\vspace{-10pt}
    \resizebox{\textwidth}{!}{
    \begin{tabular}{c c| c c c c | c c c c }
    \toprule  
      & & \multicolumn{4}{|c}{Standard CL Benchmark}&  \multicolumn{4}{|c}{Long chain of tasks}\\ 
      & & Order 1 & Order 2 & Order 3 & \textbf{Average} $(\uparrow)$ &  Order 4 & Order 5 & Order 6 & \textbf{Average} $(\uparrow)$\\
     \midrule  
     \multirow{6}{*}{\textbf{Baselines}}&
     LoRA &67.7&65.4&66.2&66.4&61.2&63.6&60.7& 61.8\\
     &EWC &72.3&65.0&70.4&69.2&59.7&61.2&65.4&62.1 \\
     &LwF &71.6&66.0&69.7&69.1&60.8&62.6&63.3&62.2 \\
     %LFPT5 &&&&&&&& \\
     &O-LoRA &78.2&76.4&\textbf{74.7}&76.5&\textbf{71.7}&73.8&70.2& 71.9\\
     
     & \textbf{DOC} (ours) &\textbf{80.5}&\textbf{78.6}&73.9&\textbf{77.7}&71.6&\textbf{74.1}&\textbf{74.4}& \textbf{73.4}\\
     
     & \textbf{DOC}-ablation&70.7&69.5&67.3&69.1&60.0&62.5&64.9&62.4 \\
    \midrule
    \multirow{4}{*}{\parbox{30pt}{\textbf{Oracle} \\ \textbf{methods}}} & 
     Replay &67.9&68.2&71.0&69.0&62.3&65.0&61.4&62.9 \\
    %ProgPrompt &&&&&&&& \\
    & PerTaskLoRA &76.9&76.9&76.9&76.9&76.8&76.8&76.8&76.8 \\
    &MTL &83.4&83.4&83.4&83.4&80.3&80.3&80.3&80.3 \\
    &ProgPrompt &77.4 &76.9 &77.9 &77.4& 76.8 &76.2 &77.1 &76.7 \\
    \bottomrule
    \end{tabular}
    }
    % %\vspace{-10pt}
    % %\vspace{-5pt}
    \label{tab:main_results_llama}
\end{table*}

\begin{table*}[!ht]
    \centering
     %\vspace{-5pt}
    \caption{Average Accuracy (AA) of different continual methods on LLaMA-13B.}
    %\vspace{-10pt}
    \resizebox{\textwidth}{!}{
    \begin{tabular}{c c| c c c c | c c c c }
    \toprule  
      & & \multicolumn{4}{|c}{Standard CL Benchmark}&  \multicolumn{4}{|c}{Long chain of tasks}\\ 
      & & Order 1 & Order 2 & Order 3 & \textbf{Average} $(\uparrow)$ &  Order 4 & Order 5 & Order 6 & \textbf{Average} $(\uparrow)$\\
     \midrule  
     \multirow{6}{*}{\textbf{Baselines}}&
     LoRA &69.2&68.0&65.7&67.6&59.9&64.7&62.0&62.2 \\
     &EWC &72.7&66.9&66.0&68.5&63.4&60.2&66.7&63.4 \\
     &LwF &71.0&70.4&72.8&71.4&64.5&62.6&65.3&64.1 \\
     %LFPT5 &&&&&&&& \\
     &O-LoRA &77.9&79.8&77.6&78.4&70.8&73.2&72.2&72.0\\
     
     & \textbf{DOC} (ours) &\textbf{79.5}&\textbf{81.2}&\textbf{79.7}&\textbf{80.1}&\textbf{72.4}&\textbf{74.0}&\textbf{76.5}& \textbf{74.3}\\
     
     & \textbf{DOC}-ablation&69.0&74.6&70.9&71.5&62.6&62.3&66.0&63.6 \\
    \midrule
    \multirow{4}{*}{\parbox{30pt}{\textbf{Oracle} \\ \textbf{methods}}} & 
     Replay &70.1&69.4&68.2&69.2&64.3&65.4&63.6&64.4 \\
    %ProgPrompt &&&&&&&& \\
    & PerTaskLoRA &77.4&77.4&77.4&77.4&78.5&78.5&78.5&78.5 \\
    &MTL &85.7&85.7&85.7&85.7&83.6&83.6&83.6&83.6 \\
    & ProgPrompt & 76.2 & 80.9 & 78.5 &78.5&79.9&80.0& 78.0 &79.3\\
    \bottomrule
    \end{tabular}
    }
    % %\vspace{-10pt}
    %\vspace{-5pt}
    \label{tab:main_results_llama_13b}

\end{table*}

\begin{table}[ht]
    \centering
    \caption{Average BWT and FWT scores of different continual methods on LLaMA-7B}
    \begin{tabular}{c|c|c|c|c}
    \toprule  
    &\multicolumn{2}{c|}{Standard CL} & \multicolumn{2}{c}{Long chain of}\\
    &\multicolumn{2}{c|}{Benchmark}&\multicolumn{2}{c}{tasks}\\
    & BWT($\uparrow$) & FWT($\uparrow$) &BWT &FWT \\
     \midrule  
     LoRA &$-14.6_{\color{red}+0.0}$&$0.6_{\color{red}+0.0}$&$-16.2_{\color{red}+0.0}$&$0.2_{\color{red}+0.0}$\\
     \midrule
     EWC &$-10.6_{\color{red}+4.0}$&$0.2_{\color{green}-0.4}$& $-14.3_{\color{red}+1.9}$ &$-1.5_{\color{green}-1.7}$ \\
     LwF &$-10.9_{\color{red}+3.7}$& $0.5_{\color{green}-0.1}$& $-15.0_{\color{red}+1.2}$ & $-0.6_{\color{green}-0.8}$\\
     %LFPT5 &&&&&&&& \\
     O-LoRA &$-1.9_{\color{red}+12.7}$& $1.4_{\color{red}+0.8}$&$-5.2_{\color{red}+11.0}$& $\mathbf{0}_{\color{green}\mathbf{-0.2}}$\\
    \textbf{DOC} (ours) &$\mathbf{-0.6}_{\mathbf{\color{red}+14.0}}$& $\mathbf{1.6}_{\color{red}\mathbf{+1.0}}$&$\mathbf{-3.4}_{\color{red}\mathbf{+12.8}}$& $-0.1_{\color{green}-0.3}$\\
    \textbf{DOC}-Ablation & $-8.8_{\color{red}+5.8}$ &$-1.5_{\color{green}-2.1}$& $-13.7_{\color{red}+2.5}$  & $-1.7_{\color{green}-1.9}$\\
    \midrule
    Replay &$-10.5_{\color{red}+4.1}$&$0_{\color{green}-0.6}$& $-14.7_{\color{red}+1.5}$&$0.2_{\color{red}+0.0}$\\
    ProgPrompt & $-0.2_{\color{red}+14.4}$ & $0.8_{\color{red}+0.2}$ & $-0.2_{\color{red}+16.0}$ & $0.1_{\color{green}-0.1}$ \\
    \bottomrule
    \end{tabular}
    % %\vspace{-10pt}
    %\vspace{-5pt}
     
    \label{tab:bwt}
\end{table}

\section{Experiments}

In this section, we test our method across various LLM continual learning benchmarks through extensive experiments to explore the practical impact on real-world continual deployment with online streaming data.

\subsection{Setup}

\noindent\textbf{Datasets and Models}. Following ProgPrompt~\citep{progressive-prompt} and O-LoRA~\citep{olora}, we employ CLBenchmark (AG News, Amazon reviews, Yelp reviews, DBpedia, Yahoo answers)~\citep{clbenchmark} to evaluate our methods, adding GLUE (MNLI, QQP, RTE, SST2)~\citep{glue},  SuperGLUE (WiC, CB, COPA, MultiRC, BoolQ)~\citep{superglue}, and IMDB review~\citep{imdb} for long-chain tasks. 
The models we use are LLaMA-7B, LLaMA-13B~\citep{llama}, and T5-Large~\citep{t5}.

\noindent\textbf{Metrics}.
Following ProgPrompt and O-LoRA, we employ \textbf{Average Accuracy (AA)} to evaluate the overall performance of continual learning, that is 
% \begin{equation}
    $\textbf{AA}(T) = \frac{1}{T}\sum_{t=1}^T a_{t,T}$
% \end{equation}
where $a_{t,T}$ is the test accuracy on the $t$-th task after fine-tuning on the $T$-th task.

In order to measure the catastrophic forgetting, we employ Backward Transfer Rate  (BWT) and Forward Transfer Rate (FWT)~\citep{CL-challenge}: 
\begin{equation}
    \textbf{BWT}(T) = \frac{1}{T-1} \sum_{t=1}^{T-1}(a_{t,T} - a_{t,t}), \quad \textbf{FWT}(T) = \frac{1}{T-1}\sum_{t=2}^T (a_{t,t} - \tilde{a}_t).
\end{equation}
Commonly, in a continual learning scenario, a negative BWT score indicates forgetting, and a negative FWT reveals that we regularize the fine-tuning process and decrease the fine-tuning performance $a_{t,t}$ compared to a standard fine-tuning performance $\tilde{a}_t$.

Overall, a regularization-based method pursues a higher BWT score representing less forgetting, at the cost of a smaller decrease in FWT score, representing less damage to the fine-tuning of each task.

\noindent\textbf{Implementation details}.
For LLaMA-7B and LLaMA-13B, we set learning rate $\alpha = \text{1e-4}$, with a batch size of 8.
For T5-Large, we let $\alpha = \text{1e-3}$ with a batch size of 64, following O-LoRA. 
Please refer to Appendix~\ref{appendix experimental details} for more details, including step number, task sequence, instructions, \textit{etc.}
%for the main experiments in 
%Table~\ref{tab:main_results_llama} and Table~\ref{tab:main_results_t5}.

\noindent\textbf{Compared methods}.
To ensure a fair comparison, we primarily focus on the recent \textbf{state-of-the-art regulation-based methods}, including EWC, LwF, and O-LoRA. We also consider fine-tuning the model with task-specific datasets sequentially using LoRA~\citep{lora}, which is a vanilla baseline and the expected lower bound of continual learning. Note that other-based methods require additional settings and are \textbf{not comparable} to ours, which is detailed in Appendix~\ref{appendix related works}. Furthermore, we present the results from several other oracle methods that are not suitable for continuous fine-tuning settings, but they can serve as upper bounds:
\begin{itemize}
    \item \textbf{Replay} replay samples from historical tasks when fine-tuning on new tasks.
    \item \textbf{PerTaskLoRA} train LoRA modules solely for each task.
    \item \textbf{MTL} train the model on all tasks as multi-task learning.
    \item \textbf{ProgPrompt}~\citep{progressive-prompt} a state-of-the-art method that updates an extending prompt in the streaming data, but task ID is required during inference.
    %, the expected upper bound of continual learning.
\end{itemize}
\begin{table*}[!ht]
    \centering
    \caption{Average Accuracy (AA) of different continual methods on T5-large}
    %\vspace{-5pt}
    \resizebox{\textwidth}{!}{
    \begin{tabular}{c c| c c c c | c c c c }
    \toprule  
    & &\multicolumn{4}{|c}{Standard CL Benchmark}&  \multicolumn{4}{|c}{Long chain of tasks}\\ 
     & &  Order 1 & Order 2 & Order 3 & Average $(\uparrow)$ &  Order 4 & Order 5 & Order 6 & Average $(\uparrow)$\\
     \midrule 
     \multirow{6}{*}{\textbf{Baselines}}
     &LoRA & 44.6 &32.7 &53.7 &43.7&2.3 &0.6 &1.9& 1.6 \\
     &EWC &48.7&47.7&54.5&50.3&45.3&44.5&45.6&45.1 \\
     &LwF &54.4&53.1&49.6&52.3&50.1&43.1&47.4&46.9 \\
     %LFPT5 &&&&&&&& \\
     &O-LoRA &75.4&75.7&\textbf{76.3}&75.8&72.3&64.8&71.6&69.6 \\

     &\textbf{DOC} (ours) &\textbf{78.8}&\textbf{78.8}&74.5&\textbf{77.4}&\textbf{72.7}&\textbf{72.4}&\textbf{74.0}&\textbf{73.0} \\
     
     &\textbf{DOC}-ablation&62.1&62.9&60.4&61.8&55.6&52.5&57.7&55.3 \\
    \midrule
    \multirow{4}{*}{\parbox{30pt}{\textbf{Oracle} \\ \textbf{methods}}}
    &Replay &55.2&56.9&61.3&57.8&55.0&54.6&53.1&54.2 \\
    %ProgPrompt&75.2&75&75.1&75.1&78.0&77.7&77.9&77.9\\
    &PerTaskLoRA&70.0&70.0&70.0&70.0&78.1&78.1&78.1&78.1 \\
    &MTL&80.0&80.0&80.0&80.0&76.5&76.5&76.5&76.5 \\

    &ProgPrompt &75.2 &75 &75.1 &75.1& 78.0 &77.7 &77.9 &77.9 \\
    \bottomrule
    \end{tabular}
    }
    %\vspace{-5pt}
    
    \label{tab:main_results_t5}
% }
\end{table*}

\subsection{Main Results}
Following ProgPrompt and O-LoRA, there are three independent runs with different task orders for different chains of tasks, as detailed in Appendix~\ref{appendix experimental details}. 
%Please refer to Appendix for details on the task sequence. 

\noindent\textbf{Overall performance}. 
The results of Average Accuracy (AA) are shown in Table~\ref{tab:main_results_llama},  Table~\ref{tab:main_results_llama_13b}, and Table~\ref{tab:main_results_t5}. 
We refer to the paper of O-LoRA~\citep{olora}, the up-to-date regularization-based method, for the results of other approaches on T5-Large, as the settings and hyperparameters of our experiments are equal. The results show that our method outperforms prior ones, especially in long-chain tasks. We respectively achieve an accuracy of 77.4 and 73.0 in the standard CL benchmark and long chains of tasks for LLaMA-7B, compared to 75.8 and 69.6 for O-LoRA, the previous state-of-the-art regularization-based method.

\noindent\textbf{Mitigating Catastrophic Forgetting}. 
The BWT and FWT results are shown in Table~\ref{tab:bwt}. The BWT score of our method is higher than that of prior approaches. We reach {-0.6} and {-3.4} for standard and long continual learning tasks, compared to \text{-1.9} and \text{-5.2} of O-LoRA, indicating that our method suffers less from catastrophic forgetting. Our FWT score, compared to other methods, indicates that we mitigate catastrophic forgetting at a slight cost to fine-tuning performance.

In summary, our method mitigates forgetting with a much higher BWT score, at the cost of a little fine-tuning performance with a slightly lower FWT score, eventually reaching effective overall performances and higher AA scores.

\subsection{Discussions}

\noindent\textbf{Computational costs}. 
The cost of storing all principal components (with maximum principal component number $K \leq 100$) is within 100MB, roughly equivalent to a few sets of LoRA modules, and is negligible compared to the cost of fine-tuning the model itself.
We employ vGPU-48GB as our device, with PyTorch 2.1.0 and CUDA 12.1, and the clock time of one training step with different regularization methods is shown in Table~\ref{time and hyperparam}. As the Online PCA technique we employ has an explicit update expression (shown in Appendix~\ref{appendix online pca}), it does not incur much extra computational costs.

\begin{table}[t]
\centering
%\vspace{-5pt}
\caption{(a) Average clock time of one fine-tuning step; (b) Average Accuracy (AA) results of standard CL Benchmark on LLaMA-7B with different LoRA rank $r$ and maximum principal component number $K$ for each new task. The results are the average of task orders 1-3.}
\resizebox{\textwidth}{!}{
\begin{tabular}{cc}
(a) & (b) \\
\begin{tabular}{c|c|c|c}
\toprule
 & LLaMA-7B & LLaMA-13B & T5-Large \\
 \midrule
 LoRA & 0.38s & 0.97s& 0.68s\\
 \midrule
 EWC &0.42s & 1.20s& 0.76s\\
 LwF & 0.40s & 1.23s& 0.76s\\
 O-LoRA & 0.42s &1.29s & 0.78s\\
 DOC(ours)& 0.46s & 1.23s& 0.80s\\
 \midrule
 ProgPrompt & 0.32s & 0.43s& 0.31s\\
 
\bottomrule
\end{tabular}
% %\vspace{-10pt}
&
\begin{tabular}{l c|c|c|c|c}
    \toprule[1pt]
    &$K$&$32$&48&$64$&$96$\\
     \midrule
     %\multirow{3}{*}{\parbox{30pt}{\textbf{Oracle} \\ \textbf{methods}}}
     \multirow{2}{*}{\parbox{30pt}
     {\textbf{DOC} \\ (ours)}} & $r=16$& 76.1&77.6&76.5& 76.5\\ 
     &$r=64$ &77.9&78.5&78.4& \textbf{78.7}\\
    \midrule[1pt]
    % \midrule
    \multirow{2}{*}{LoRA}
     & $r=16$&\multicolumn{4}{c}{65.4}\\
     & $r=64$&\multicolumn{4}{c}{67.4}\\ 
     \midrule
     \multirow{2}{*}{O-LoRA}
     & $r=16$&\multicolumn{4}{c}{77.0}\\
     & $r=64$&\multicolumn{4}{c}{76.8}\\ 
    \bottomrule[1pt]
    \end{tabular}

\end{tabular}
}
\label{time and hyperparam}
%\vspace{-10pt}
\end{table}

\noindent\textbf{The choice of hyperparameters}.
We present the following empirical study regarding different choices of LoRA rank, say $r$, for fine-tuning,  and the maximum principal component number for each new task, say $K$, for functional direction tracking. The results are shown in Table~\ref{time and hyperparam}. Overall, adequate principal components cooperating with higher LoRA ranks are able to cover and protect more critical functional directions for historical tasks, thus ensuring a better historical functional preservation and task accuracy. The results show that the variation between different choices of hyperparameters is little, revealing the robustness of our method.

\noindent\textbf{Ablation study}.
We conduct a trial on freezing the update of principal components to investigate the impact of functional direction tracking. Specifically, we cease updating principal components after their initialization during the first 10\% fine-tuning steps for each task. The results are shown in the \textbf{DOC-ablation} line in Table~\ref{tab:main_results_llama},\ref{tab:main_results_llama_13b}, \ref{tab:bwt}, and~\ref{tab:main_results_t5}. The decrease in continual learning performance in the ablation experiment indicates that it is tracking the functional directions that mitigate catastrophic forgetting and enhance the performance of continual learning.

\section{Conclusion}

In this paper, we introduce a novel regularization-based approach that leverages functional direction tracking for continual learning in language models. We identify that the drift of functional directions is the key issue for regularization-based continual learning approaches, and the proposed method systematically addresses the drift issue by updating the functional directions dynamically with Online PCA during the fine-tuning process. Empirical evaluations verify the effectiveness of our tracking method and underscore its efficacy in enhancing continual learning performance. For limitations and future directions, please refer to Appendix~\ref{limitations}~and~\ref{future directions}.

\section*{Ethics statement}
This work focuses on developing fine-tuning methods to mitigate catastrophic forgetting in LLMs for continual learning, with no involvement of human subjects, sensitive personal data, or high-risk real-world deployments.  

\section*{Reproducibility statement}

Our code will be available upon publication. All datasets and LLMs we used in experiments are publicly available online.

% \clearpage
\bibliography{ref}

\begin{thebibliography}{44}
\providecommand{\natexlab}[1]{#1}
\providecommand{\url}[1]{\texttt{#1}}
\expandafter\ifx\csname urlstyle\endcsname\relax
  \providecommand{\doi}[1]{doi: #1}\else
  \providecommand{\doi}{doi: \begingroup \urlstyle{rm}\Url}\fi

\bibitem[Aghajanyan et~al.(2020)Aghajanyan, Zettlemoyer, and Gupta]{intrinsiclora}
Armen Aghajanyan, Luke Zettlemoyer, and Sonal Gupta.
\newblock Intrinsic dimensionality explains the effectiveness of language model fine-tuning, 2020.
\newblock URL \url{https://arxiv.org/abs/2012.13255}.

\bibitem[Arora et~al.(2012)Arora, Cotter, Livescu, and Srebro]{incrementalpca1}
Raman Arora, Andrew Cotter, Karen Livescu, and Nathan Srebro.
\newblock Stochastic optimization for pca and pls.
\newblock In \emph{2012 50th Annual Allerton Conference on Communication, Control, and Computing (Allerton)}, pp.\  861--868, 2012.
\newblock \doi{10.1109/Allerton.2012.6483308}.

\bibitem[Black et~al.(2022)Black, Sharkey, Grinsztajn, Winsor, Braun, Merizian, Parker, Guevara, Millidge, Alfour, and Leahy]{polytope}
Sid Black, Lee Sharkey, Leo Grinsztajn, Eric Winsor, Dan Braun, Jacob Merizian, Kip Parker, Carlos~Ramón Guevara, Beren Millidge, Gabriel Alfour, and Connor Leahy.
\newblock Interpreting neural networks through the polytope lens, 2022.
\newblock URL \url{https://arxiv.org/abs/2211.12312}.

\bibitem[Cardot \& Degras(2015)Cardot and Degras]{pcasurvey}
Hervé Cardot and David Degras.
\newblock Online principal component analysis in high dimension: Which algorithm to choose?, 2015.
\newblock URL \url{https://arxiv.org/abs/1511.03688}.

\bibitem[de~Masson~d'Autume et~al.(2019)de~Masson~d'Autume, Ruder, Kong, and Yogatama]{rehearsal1}
Cyprien de~Masson~d'Autume, Sebastian Ruder, Lingpeng Kong, and Dani Yogatama.
\newblock Episodic memory in lifelong language learning, 2019.
\newblock URL \url{https://arxiv.org/abs/1906.01076}.

\bibitem[Farajtabar et~al.(2019)Farajtabar, Azizan, Mott, and Li]{ogd}
Mehrdad Farajtabar, Navid Azizan, Alex Mott, and Ang Li.
\newblock Orthogonal gradient descent for continual learning, 2019.
\newblock URL \url{https://arxiv.org/abs/1910.07104}.

\bibitem[Hendrycks et~al.(2021{\natexlab{a}})Hendrycks, Burns, Basart, Critch, Li, Song, and Steinhardt]{ethics}
Dan Hendrycks, Collin Burns, Steven Basart, Andrew Critch, Jerry Li, Dawn Song, and Jacob Steinhardt.
\newblock Aligning ai with shared human values.
\newblock \emph{Proceedings of the International Conference on Learning Representations (ICLR)}, 2021{\natexlab{a}}.

\bibitem[Hendrycks et~al.(2021{\natexlab{b}})Hendrycks, Burns, Basart, Zou, Mazeika, Song, and Steinhardt]{mmlu}
Dan Hendrycks, Collin Burns, Steven Basart, Andy Zou, Mantas Mazeika, Dawn Song, and Jacob Steinhardt.
\newblock Measuring massive multitask language understanding.
\newblock \emph{Proceedings of the International Conference on Learning Representations (ICLR)}, 2021{\natexlab{b}}.

\bibitem[Hu et~al.(2021)Hu, Shen, Wallis, Allen-Zhu, Li, Wang, Wang, and Chen]{lora}
Edward~J. Hu, Yelong Shen, Phillip Wallis, Zeyuan Allen-Zhu, Yuanzhi Li, Shean Wang, Lu~Wang, and Weizhu Chen.
\newblock Lora: Low-rank adaptation of large language models, 2021.
\newblock URL \url{https://arxiv.org/abs/2106.09685}.

\bibitem[Huang et~al.(2021)Huang, Zhang, Chen, Wang, and Yang]{rehearsal3}
Yufan Huang, Yanzhe Zhang, Jiaao Chen, Xuezhi Wang, and Diyi Yang.
\newblock Continual learning for text classification with information disentanglement based regularization, 2021.
\newblock URL \url{https://arxiv.org/abs/2104.05489}.

\bibitem[Jang et~al.(2023)Jang, Kim, Ye, Kim, Logeswaran, Lee, Lee, and Seo]{expert1}
Joel Jang, Seungone Kim, Seonghyeon Ye, Doyoung Kim, Lajanugen Logeswaran, Moontae Lee, Kyungjae Lee, and Minjoon Seo.
\newblock Exploring the benefits of training expert language models over instruction tuning, 2023.
\newblock URL \url{https://arxiv.org/abs/2302.03202}.

\bibitem[Kirkpatrick et~al.(2017)Kirkpatrick, Pascanu, Rabinowitz, Veness, Desjardins, Rusu, Milan, Quan, Ramalho, Grabska-Barwinska, Hassabis, Clopath, Kumaran, and Hadsell]{ewc}
James Kirkpatrick, Razvan Pascanu, Neil Rabinowitz, Joel Veness, Guillaume Desjardins, Andrei~A. Rusu, Kieran Milan, John Quan, Tiago Ramalho, Agnieszka Grabska-Barwinska, Demis Hassabis, Claudia Clopath, Dharshan Kumaran, and Raia Hadsell.
\newblock Overcoming catastrophic forgetting in neural networks.
\newblock \emph{Proceedings of the National Academy of Sciences}, 114\penalty0 (13):\penalty0 3521–3526, March 2017.
\newblock ISSN 1091-6490.
\newblock \doi{10.1073/pnas.1611835114}.
\newblock URL \url{http://dx.doi.org/10.1073/pnas.1611835114}.

\bibitem[Krasulina(1970)]{stochasticpca2}
T.~P. Krasulina.
\newblock Method of stochastic approximation in the determination of the largest eigenvalue of the mathematical expectation of random matrices.
\newblock \emph{Automation and Remote Control}, pp.\  215--221, 1970.
\newblock Originally published in \emph{Avtomatika i Telemekhanika}, 1970, no. 2, pp. 50--56.

\bibitem[Levey \& Lindenbaum(2000)Levey and Lindenbaum]{incrementalpca2}
A.~Levey and M.~Lindenbaum.
\newblock Sequential karhunen-loeve basis extraction and its application to images.
\newblock \emph{IEEE Transactions on Image Processing}, 9\penalty0 (8):\penalty0 1371--1374, 2000.
\newblock \doi{10.1109/83.855432}.

\bibitem[Li \& Hoiem(2017)Li and Hoiem]{lwf}
Zhizhong Li and Derek Hoiem.
\newblock Learning without forgetting, 2017.
\newblock URL \url{https://arxiv.org/abs/1606.09282}.

\bibitem[Maas et~al.(2011)Maas, Daly, Pham, Huang, Ng, and Potts]{imdb}
Andrew~L. Maas, Raymond~E. Daly, Peter~T. Pham, Dan Huang, Andrew~Y. Ng, and Christopher Potts.
\newblock Learning word vectors for sentiment analysis.
\newblock In Dekang Lin, Yuji Matsumoto, and Rada Mihalcea (eds.), \emph{Proceedings of the 49th Annual Meeting of the Association for Computational Linguistics: Human Language Technologies}, pp.\  142--150, Portland, Oregon, USA, June 2011. Association for Computational Linguistics.
\newblock URL \url{https://aclanthology.org/P11-1015/}.

\bibitem[Michaud et~al.(2024)Michaud, Liu, Girit, and Tegmark]{quantom}
Eric~J. Michaud, Ziming Liu, Uzay Girit, and Max Tegmark.
\newblock The quantization model of neural scaling, 2024.
\newblock URL \url{https://arxiv.org/abs/2303.13506}.

\bibitem[Mok et~al.(2023)Mok, Do, Lee, Taghavi, Yu, and Yoon]{rehearsal2}
Jisoo Mok, Jaeyoung Do, Sungjin Lee, Tara Taghavi, Seunghak Yu, and Sungroh Yoon.
\newblock Large-scale lifelong learning of in-context instructions and how to tackle it.
\newblock In Anna Rogers, Jordan Boyd-Graber, and Naoaki Okazaki (eds.), \emph{Proceedings of the 61st Annual Meeting of the Association for Computational Linguistics (Volume 1: Long Papers)}, pp.\  12573--12589, Toronto, Canada, July 2023. Association for Computational Linguistics.
\newblock \doi{10.18653/v1/2023.acl-long.703}.
\newblock URL \url{https://aclanthology.org/2023.acl-long.703/}.

\bibitem[Oja(1992)]{stochasticpca4}
Erkki Oja.
\newblock Principal components, minor components, and linear neural networks.
\newblock \emph{Neural Networks}, 5\penalty0 (6):\penalty0 927--935, 1992.
\newblock ISSN 0893-6080.
\newblock \doi{https://doi.org/10.1016/S0893-6080(05)80089-9}.
\newblock URL \url{https://www.sciencedirect.com/science/article/pii/S0893608005800899}.

\bibitem[Oja \& Karhunen(1985)Oja and Karhunen]{stochasticpca3}
Erkki Oja and Juha Karhunen.
\newblock On stochastic approximation of the eigenvectors and eigenvalues of the expectation of a random matrix.
\newblock \emph{Journal of Mathematical Analysis and Applications}, 106\penalty0 (1):\penalty0 69--84, 1985.
\newblock ISSN 0022-247X.
\newblock \doi{https://doi.org/10.1016/0022-247X(85)90131-3}.
\newblock URL \url{https://www.sciencedirect.com/science/article/pii/0022247X85901313}.

\bibitem[Olah et~al.(2018)Olah, Satyanarayan, Johnson, Carter, Schubert, Ye, and Mordvintsev]{direction2}
Chris Olah, Arvind Satyanarayan, Ian Johnson, Shan Carter, Ludwig Schubert, Katherine Ye, and Alexander Mordvintsev.
\newblock The building blocks of interpretability.
\newblock \emph{Distill}, 2018.
\newblock \doi{10.23915/distill.00010}.
\newblock https://distill.pub/2018/building-blocks.

\bibitem[Olah et~al.(2020)Olah, Cammarata, Schubert, Goh, Petrov, and Carter]{direction1}
Chris Olah, Nick Cammarata, Ludwig Schubert, Gabriel Goh, Michael Petrov, and Shan Carter.
\newblock Zoom in: An introduction to circuits.
\newblock \emph{Distill}, 2020.
\newblock \doi{10.23915/distill.00024.001}.
\newblock https://distill.pub/2020/circuits/zoom-in.

\bibitem[Peng et~al.(2024)Peng, Tian, Liu, Yang, and Jia]{expertavg}
Bohao Peng, Zhuotao Tian, Shu Liu, Mingchang Yang, and Jiaya Jia.
\newblock Scalable language model with generalized continual learning, 2024.
\newblock URL \url{https://arxiv.org/abs/2404.07470}.

\bibitem[Qin \& Joty(2022)Qin and Joty]{lfpt5}
Chengwei Qin and Shafiq Joty.
\newblock Lfpt5: A unified framework for lifelong few-shot language learning based on prompt tuning of t5, 2022.
\newblock URL \url{https://arxiv.org/abs/2110.07298}.

\bibitem[Raffel et~al.(2023)Raffel, Shazeer, Roberts, Lee, Narang, Matena, Zhou, Li, and Liu]{t5}
Colin Raffel, Noam Shazeer, Adam Roberts, Katherine Lee, Sharan Narang, Michael Matena, Yanqi Zhou, Wei Li, and Peter~J. Liu.
\newblock Exploring the limits of transfer learning with a unified text-to-text transformer, 2023.
\newblock URL \url{https://arxiv.org/abs/1910.10683}.

\bibitem[Razdaibiedina et~al.(2023)Razdaibiedina, Mao, Hou, Khabsa, Lewis, and Almahairi]{progressive-prompt}
Anastasia Razdaibiedina, Yuning Mao, Rui Hou, Madian Khabsa, Mike Lewis, and Amjad Almahairi.
\newblock Progressive prompts: Continual learning for language models, 2023.
\newblock URL \url{https://arxiv.org/abs/2301.12314}.

\bibitem[Sanger(1989)]{stochasitcpca1}
Terence~D. Sanger.
\newblock Optimal unsupervised learning in a single-layer linear feedforward neural network.
\newblock \emph{Neural Networks}, 2\penalty0 (6):\penalty0 459--473, 1989.
\newblock ISSN 0893-6080.
\newblock \doi{https://doi.org/10.1016/0893-6080(89)90044-0}.
\newblock URL \url{https://www.sciencedirect.com/science/article/pii/0893608089900440}.

\bibitem[Saxena \& Cunningham(2019)Saxena and Cunningham]{direction3}
Shreya Saxena and John~P Cunningham.
\newblock Towards the neural population doctrine.
\newblock \emph{Current Opinion in Neurobiology}, 55:\penalty0 103--111, 2019.
\newblock ISSN 0959-4388.
\newblock \doi{https://doi.org/10.1016/j.conb.2019.02.002}.
\newblock URL \url{https://www.sciencedirect.com/science/article/pii/S0959438818300990}.
\newblock Machine Learning, Big Data, and Neuroscience.

\bibitem[Taori et~al.(2023)Taori, Gulrajani, Zhang, Dubois, Li, Guestrin, Liang, and Hashimoto]{alpaca}
Rohan Taori, Ishaan Gulrajani, Tianyi Zhang, Yann Dubois, Xuechen Li, Carlos Guestrin, Percy Liang, and Tatsunori~B. Hashimoto.
\newblock Stanford alpaca: An instruction-following llama model.
\newblock \url{https://github.com/tatsu-lab/stanford_alpaca}, 2023.

\bibitem[Touvron et~al.(2023)Touvron, Martin, Stone, Albert, Almahairi, Babaei, Bashlykov, Batra, Bhargava, Bhosale, et~al.]{llama}
Hugo Touvron, Louis Martin, Kevin Stone, Peter Albert, Amjad Almahairi, Yasmine Babaei, Nikolay Bashlykov, Soumya Batra, Prajjwal Bhargava, Shruti Bhosale, et~al.
\newblock Llama 2: Open foundation and fine-tuned chat models.
\newblock \emph{arXiv preprint arXiv:2307.09288}, 2023.

\bibitem[Wang et~al.(2019)Wang, Singh, Michael, Hill, Levy, and Bowman]{glue}
Alex Wang, Amanpreet Singh, Julian Michael, Felix Hill, Omer Levy, and Samuel~R. Bowman.
\newblock Glue: A multi-task benchmark and analysis platform for natural language understanding, 2019.
\newblock URL \url{https://arxiv.org/abs/1804.07461}.

\bibitem[Wang et~al.(2020)Wang, Pruksachatkun, Nangia, Singh, Michael, Hill, Levy, and Bowman]{superglue}
Alex Wang, Yada Pruksachatkun, Nikita Nangia, Amanpreet Singh, Julian Michael, Felix Hill, Omer Levy, and Samuel~R. Bowman.
\newblock Superglue: A stickier benchmark for general-purpose language understanding systems, 2020.
\newblock URL \url{https://arxiv.org/abs/1905.00537}.

\bibitem[Wang et~al.(2024)Wang, Adel, Lange, Strötgen, and Schütze]{expert2}
Mingyang Wang, Heike Adel, Lukas Lange, Jannik Strötgen, and Hinrich Schütze.
\newblock Rehearsal-free modular and compositional continual learning for language models, 2024.
\newblock URL \url{https://arxiv.org/abs/2404.00790}.

\bibitem[Wang et~al.(2023{\natexlab{a}})Wang, Chen, Ge, Xia, Bao, Zheng, Zhang, Gui, and Huang]{olora}
Xiao Wang, Tianze Chen, Qiming Ge, Han Xia, Rong Bao, Rui Zheng, Qi~Zhang, Tao Gui, and Xuanjing Huang.
\newblock Orthogonal subspace learning for language model continual learning, 2023{\natexlab{a}}.
\newblock URL \url{https://arxiv.org/abs/2310.14152}.

\bibitem[Wang et~al.(2023{\natexlab{b}})Wang, Liu, Ji, Wang, Wu, Jiang, Chao, Han, Wang, Shao, and Zeng]{expert4}
Zhicheng Wang, Yufang Liu, Tao Ji, Xiaoling Wang, Yuanbin Wu, Congcong Jiang, Ye~Chao, Zhencong Han, Ling Wang, Xu~Shao, and Wenqiu Zeng.
\newblock Rehearsal-free continual language learning via efficient parameter isolation.
\newblock In Anna Rogers, Jordan Boyd-Graber, and Naoaki Okazaki (eds.), \emph{Proceedings of the 61st Annual Meeting of the Association for Computational Linguistics (Volume 1: Long Papers)}, pp.\  10933--10946, Toronto, Canada, July 2023{\natexlab{b}}. Association for Computational Linguistics.
\newblock \doi{10.18653/v1/2023.acl-long.612}.
\newblock URL \url{https://aclanthology.org/2023.acl-long.612/}.

\bibitem[Wang et~al.(2022)Wang, Zhang, Lee, Zhang, Sun, Ren, Su, Perot, Dy, and Pfister]{l2p}
Zifeng Wang, Zizhao Zhang, Chen-Yu Lee, Han Zhang, Ruoxi Sun, Xiaoqi Ren, Guolong Su, Vincent Perot, Jennifer Dy, and Tomas Pfister.
\newblock Learning to prompt for continual learning, 2022.
\newblock URL \url{https://arxiv.org/abs/2112.08654}.

\bibitem[Weng et~al.(2003)Weng, Zhang, and Hwang]{ccipca}
Juyang Weng, Yilu Zhang, and Wey-Shiuan Hwang.
\newblock Candid covariance-free incremental principal component analysis.
\newblock \emph{IEEE Transactions on Pattern Analysis and Machine Intelligence}, 25\penalty0 (8):\penalty0 1034--1040, 2003.
\newblock \doi{10.1109/TPAMI.2003.1217609}.

\bibitem[Wu et~al.(2024{\natexlab{a}})Wu, Gan, Ge, Lu, Wang, Feng, Shan, and Luo]{llama-pro}
Chengyue Wu, Yukang Gan, Yixiao Ge, Zeyu Lu, Jiahao Wang, Ye~Feng, Ying Shan, and Ping Luo.
\newblock Llama pro: Progressive llama with block expansion, 2024{\natexlab{a}}.
\newblock URL \url{https://arxiv.org/abs/2401.02415}.

\bibitem[Wu et~al.(2022)Wu, Caccia, Li, Li, Qi, and Haffari]{CL-challenge}
Tongtong Wu, Massimo Caccia, Zhuang Li, Yuan-Fang Li, Guilin Qi, and Gholamreza Haffari.
\newblock Pretrained language model in continual learning: A comparative study.
\newblock In \emph{International Conference on Learning Representations}, 2022.
\newblock URL \url{https://openreview.net/forum?id=figzpGMrdD}.

\bibitem[Wu et~al.(2024{\natexlab{b}})Wu, Luo, Li, Pan, Vu, and Haffari]{llmsurvey}
Tongtong Wu, Linhao Luo, Yuan-Fang Li, Shirui Pan, Thuy-Trang Vu, and Gholamreza Haffari.
\newblock Continual learning for large language models: A survey, 2024{\natexlab{b}}.
\newblock URL \url{https://arxiv.org/abs/2402.01364}.

\bibitem[Zenke et~al.(2017)Zenke, Poole, and Ganguli]{synapsis}
Friedemann Zenke, Ben Poole, and Surya Ganguli.
\newblock Continual learning through synaptic intelligence, 2017.
\newblock URL \url{https://arxiv.org/abs/1703.04200}.

\bibitem[Zhang et~al.(2016)Zhang, Zhao, and LeCun]{clbenchmark}
Xiang Zhang, Junbo Zhao, and Yann LeCun.
\newblock Character-level convolutional networks for text classification, 2016.
\newblock URL \url{https://arxiv.org/abs/1509.01626}.

\bibitem[Zhang \& Weng(2001)Zhang and Weng]{ccipca-convergence}
Yilu Zhang and Juyang Weng.
\newblock Convergence analysis of complementary candid incremental principal component analysis.
\newblock 2001.

\bibitem[Zheng et~al.(2024)Zheng, Qiu, Shi, and Ma]{lifelongllmsurvey}
Junhao Zheng, Shengjie Qiu, Chengming Shi, and Qianli Ma.
\newblock Towards lifelong learning of large language models: A survey, 2024.
\newblock URL \url{https://arxiv.org/abs/2406.06391}.

\end{thebibliography}
\bibliographystyle{iclr2026_conference}

\clearpage
\appendix

\section{Online PCA in our method}~\label{appendix online pca}
In our method, we extract the basis of functional directions from their linear combinations using Online PCA~\citep{pcasurvey}. Specifically, we employ a modified Candid Covariance-
free Incremental PCA (CCIPCA)~\citep{ccipca} to implement Online PCA. 

\noindent\textbf{The CCIPCA technique}.
Let $\Gamma = \frac{1}{T-1}HH^\top$ be the covariance matrix $H=(h_1, h_2,...,h_N)$  with standardized datas $h_1, h_2,...,h_N$. Recall the goal of the PCA is to find the eigenvector $u$ and the eigenvalue $\lambda$ of $\Gamma$ that satisfy  
\begin{equation}\label{ccipca:target}
    \Gamma u = \lambda u.
\end{equation}
The idea of CCIPCA is as follows. For the first eigenvector $v^1$, assume that estimates $v^1_0,...,v_{T-1}^1$ of $v = \lambda u$ have been constructed in previous steps $t=1,2,...,T-1$. We substitute $h_th_t^\top$ to $\Gamma$ and $\frac{v_{t-1}^1}{\Vert v_{t-1}^1 \Vert}$ to $u$ in the eigenequation~\eqref{ccipca:target} for $t = 1,...,T$, and average the results:
\begin{equation}\label{ccipca:1}
    v_T^1 = \frac{1}{T}\sum_{t=1}^T h_t h_t^\top 
    \frac{v_{t-1}^1}{\Vert v_{t-1}^1 \Vert}.
\end{equation}
Note that CCIPCA requires no historical datas $\{ h_1, h_2,...,h_{T-1}\}$ , as equation~\eqref{ccipca:1} can be conveniently written in recursive form as:
\begin{equation}\label{ccipca:2}
    v_{T+1}^1 = \frac{T-l}{T+1} v_T^1 + 
    \frac{1+l}{T+1} h_{T+1} h_{T+1}^\top 
    \frac{v_{T}^1}{\Vert v_{T}^1 \Vert},
\end{equation}
where an amnesic factor $l \geq 0$ is introduced to handle nonstationary data generation, and the initialization is $v_0 = h_1$. The almost-sure convergence of equation~\eqref{ccipca:2} has been proved by \citep{ccipca-convergence}. For estimating more than one eigenvector, say $v^1, v^2, ...,v^K$, to update the $K$-th eigenvector $v^K_{T+1}$, simply replace the input vector $h_{T+1}$ in equation~\eqref{ccipca:2} with the following residual cutting:
\begin{equation}\label{ccipca-residual}
    h_{T+1}^* = h_{T+1} - \sum_{k=1}^{K-1}
    \frac{h_{T+1} \cdot v^k_{T}}{\Vert v_T^k \Vert ^2} \cdot v^k_{T}.
\end{equation}

\noindent\textbf{Modified CCIPCA for tracking functional directions}.
To deal with the issue of functional direction drift, we introduce a tracking factor $\epsilon \in (0,1)$ to equation~\eqref{ccipca:2} for a faster update:

\begin{equation}\label{ccipca:4}
    v_{T+1}^1 = \eta \cdot v_T^1 + 
    (1-\eta) \cdot h_{T+1} h_{T+1}^\top
    \frac{v_{T}^1}{\Vert v_{T}^1 \Vert} ,
\end{equation}
where $\eta =\frac{T-l}{T+1}\cdot(1-\epsilon)$. Note that the convergence of equation~\eqref{ccipca:4} is disturbed for tracking. Algorithm \ref{ccipca-algorithm} summarizes the modified tracking CCIPCA method, which additionally employs a residual threshold $\delta \in (0,1)$ to append new components automatically (lines 7-10).

\begin{figure}[t]
    \centering
    \begin{tabular}{c}
    \includegraphics[width=0.8\textwidth]{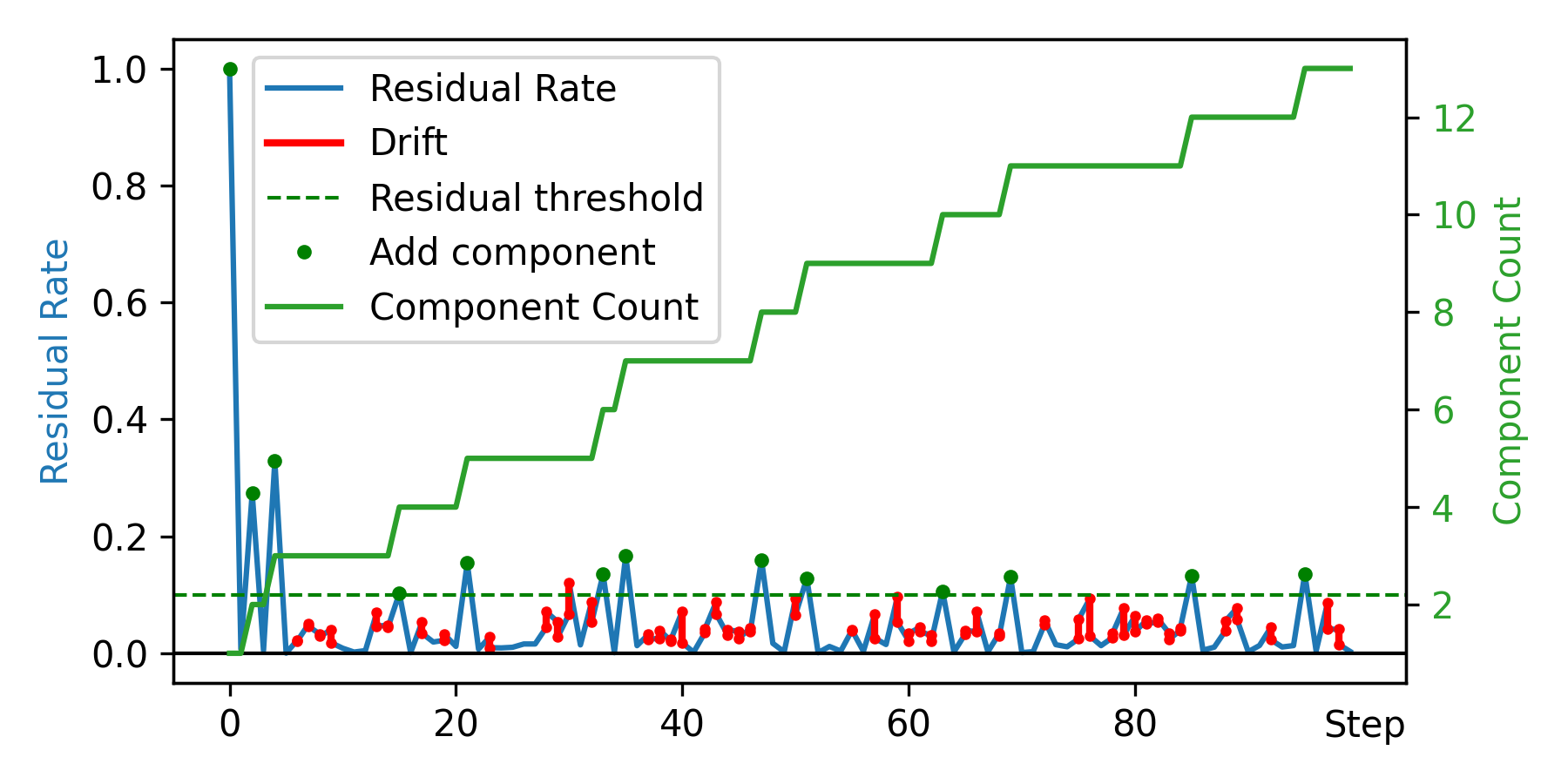} \\
    \end{tabular}
    \vspace{-10pt}
    \caption{The update of principal components. If the residual rate is over the threshold, we add a new component for it. 
    %as presented in Algorithm~\ref{ccipca-algorithm}. 
    The red line shows that we track the drift by adjusting the tracking factor $\epsilon$, whose increase mostly reduces the residual rate for better drift tracking.
    }
    \vspace{-10pt}
    \label{fig-ccipca-process}
\end{figure}

\noindent\textbf{An example of functional direction tracking}.
We present the following example on the working process of functional direction tracking as a reference for Algorithm~\ref{ccipca-algorithm}. Still, we take fine-tuning Llama-2 on CLBenchmark as an example. 
As shown in Figure~\ref{fig-ccipca-process}, we update the principal components based on the residual rate $\frac{\Vert h^*_t \Vert}{\Vert h_t \Vert}$ with residual threshold $\delta$, abiding by lines 7-10 in Algorithm~\ref{ccipca-algorithm}.
Note that a lower residual rate indicates more complete coverage of LoRA increment with existing principal components. The tracking factor $\epsilon$ is adjusted dynamically following the increase and decrease of the residual rate, which is executed by redoing lines 2-6 in Algorithm~\ref{ccipca-algorithm} with adjusted $\epsilon$ and $\eta$. 
The results show that we continuously keep the residual rate less than 10\%, covering 90\% of the LoRA increment.

\begin{algorithm}[tb]
\caption{Online PCA for Tracking Functional Directions}
\label{ccipca-algorithm}
\textbf{Parameter}: Maximum principal component number $K_\text{max}$, amnesic factor $l$, tracking factor $\epsilon$, residual threshold $\delta$ \\
\textbf{Initialization}: current principal component number $n=0$\\
\textbf{Input}: The incoming model state data $h_{T+1}$\\
\textbf{Output}: Updated principal components 
$v^1_{T+1},...,v_{T+1}^K$\\
\begin{algorithmic}[1] %[1] enables line numbers
\STATE residual $h_{T+1}^* = h_{T+1}$
\STATE $\eta =\frac{T-l}{T+1}\cdot(1-\epsilon)$
\FOR{$k$ in $range(n)$}
\STATE update $v^k_{T+1}$ using $h^*_{T+1}$ with equation~\eqref{ccipca:4}
\STATE update $h_{T+1}^*$ with equation~\eqref{ccipca-residual}
\ENDFOR
\IF{$n<K_\text{max}$ and $\frac{\Vert h^*_{T+1} \Vert}{\Vert h_{T+1} \Vert} > \delta$}
\STATE add a new component $v^{n+1}_{T+1} = h^*_{T+1}$
\STATE $n = n+1$
\ENDIF
\end{algorithmic}
\end{algorithm}
%\vspace{-20pt}

\section{Addtional experimental details}\label{appendix experimental details}
For the Online PCA above, we let the amnestic factor $l=2$, following the recommendation of~\citep{ccipca}. The empirical value of the tracking factor is that $\epsilon \in (0, 0.1)$, which is adjusted with the increase and decrease of the residual rate, and the residual threshold $\delta = 0.1$ for adding components automatically. For each new task, we enhance the maximum principal component number $K_\text{max}$ by 48. 

We follow O-LoRA\citep{olora} and Progressive Prompt~\citep{progressive-prompt} for the following continual learning settings:

\textbf{Dataset details}.
Table~\ref{tab:dataset} shows details of the datasets we employ for continual learning experiments, along with their evaluation metrics. Overall, we used datasets from CL benchmark~\citep{clbenchmark}, GLUE~\citep{glue}, and SuperGLUE~\citep{superglue} benchmarks, adding the IMDB movie reviews~\citep{imdb}. We randomly sample 100-10000 samples for each dataset, depending on their size, and fine-tune for 1000 steps for each incoming dataset in streaming data.

\textbf{Task sequence of continual learning}.
The task orders used for our CL experiments across LLaMA and T5 models are shown in Table~\ref{tab:order}.

\textbf{Prompts for different tasks}.
Table~\ref{tab:prompt} shows prompts for different tasks. NLI denotes natural language inference, including MNLI, RTE, CB. SC denotes sentiment analysis, including Amazon, Yelp, SST-2, IMDB. TC denotes topic classification, including AG News, DBpedia, Yahoo. 
%\subsection{Method insights}

\begin{table*}[t]
\centering
\caption{The details of 15 datasets used in the CL experiments, following O-LoRA and Progressive Prompt. NLI denotes natural language
inference, QA denotes the question and answer task.}
\begin{tabular}{l|llll}
\toprule
\textbf{Dataset name} & \textbf{Category} & \textbf{Task}             & \textbf{Domain}     & \textbf{Metric} \\ \midrule
1. Yelp               & CL Benchmark      & sentiment analysis        & Yelp reviews        & accuracy        \\
2. Amazon             & CL Benchmark      & sentiment analysis        & Amazon reviews      & accuracy        \\
3. DBpedia            & CL Benchmark      & topic classification      & Wikipedia           & accuracy        \\
4. Yahoo              & CL Benchmark      & topic classification      & Yahoo Q\&A          & accuracy        \\
5. AG News            & CL Benchmark      & topic classification      & news                & accuracy        \\
6. MNLI               & GLUE              & NLI                       & various             & accuracy        \\
7. QQP                & GLUE              & paragraph detection       & Quora               & accuracy        \\
8. RTE                & GLUE              & NLI                       & news, Wikipedia     & accuracy        \\
9. SST-2              & GLUE              & sentiment analysis        & movie reviews       & accuracy        \\
10. WiC               & SuperGLUE         & word sense disambiguation & lexical databases   & accuracy        \\
11. CB                & SuperGLUE         & NLI                       & various             & accuracy        \\
12. COPA              & SuperGLUE         & QA                        & blogs, encyclopedia & accuracy        \\
13. BoolQA            & SuperGLUE         & boolean QA                & Wikipedia           & accuracy        \\
14. MultiRC           & SuperGLUE         & QA                        & various             & accuracy        \\
15. IMDB              & SuperGLUE         & sentiment analysis        & movie reviews       & accuracy        \\ \bottomrule
\end{tabular}

\label{tab:dataset}
\end{table*}

\begin{table*}[t]
\centering
\caption{Different orders of task sequences used for continual learning experiments. Orders
1-3 correspond to the standard CL benchmark,  orders 4-6 are long chain of tasks, following O-LoRA and Progressive Prompt.}
\begin{tabular}{cll}
\toprule
\textbf{Order} & \textbf{Task Sequence}                                                                                                                                \\ 
\midrule
1                   & dbpedia → amazon → yahoo → ag                                                                                                                         \\
2                  & dbpedia → amazon → ag → yahoo                                                                                                                         \\
3                   & yahoo → amazon → ag → dbpedia                                                                                                                         \\ \midrule
4                      & \begin{tabular}[c]{@{}l@{}}mnli → cb → wic → copa → qqp → boolqa → rte → imdb →\\ yelp → amazon → sst-2 → dbpedia → ag → multirc → yahoo\end{tabular} \\
5                      & \begin{tabular}[c]{@{}l@{}}multirc → boolqa → wic → mnli → cb → copa → qqp → rte\\ → imdb → sst-2 → dbpedia → ag → yelp → amazon → yahoo\end{tabular} \\
6                      & \begin{tabular}[c]{@{}l@{}}yelp → amazon → mnli → cb → copa → qqp → rte → imdb →\\ sst-2 → dbpedia → ag → yahoo → multirc → boolqa → wic\end{tabular} \\ \bottomrule
\end{tabular}

\label{tab:order}
\vspace{-10pt}
\end{table*}

\begin{table*}[ht]

\centering
\caption{Instructions for different tasks, following O-LoRA and Progressive Prompt.}
\begin{tabular}{cl}
\toprule
\textbf{Task}                                                       & \multicolumn{1}{c}{\textbf{Prompts}}                                                                                                                                  \\ \midrule
NLI                                                                 & \begin{tabular}[c]{@{}l@{}}What is the logical relationship between the "sentence 1" and the "sentence 2"? \\ Choose one from the options.\end{tabular}                \\ \midrule
QQP                                                                 & \begin{tabular}[c]{@{}l@{}}Whether the "first sentence" and the "second sentence" have the same meaning? \\ Choose one from the options.\end{tabular}                  \\ \midrule
\begin{tabular}[c]{@{}c@{}}SC\end{tabular}   & What is the sentiment of the following paragraph? Choose one from the options.                                                                                         \\ \midrule
\begin{tabular}[c]{@{}c@{}}TC\end{tabular} & What is the topic of the following paragraph? Choose one from the options.                                                                                             \\ \midrule
BoolQA                                                              & \begin{tabular}[c]{@{}l@{}}According to the following passage, is the question true or false? Choose one \\ from the options.\end{tabular}                             \\ \midrule
MultiRC                                                             & \begin{tabular}[c]{@{}l@{}}According to the following passage and question, is the candidate answer true \\ or false? Choose one from the options.\end{tabular}        \\ \midrule
WiC                                                                 & \begin{tabular}[c]{@{}l@{}}Given a word and two sentences, whether the word is used with the same sense \\ in both sentences? Choose one from the options.\end{tabular} \\ \bottomrule
\end{tabular}

\label{tab:prompt}
\end{table*}

\section{Details for computation}\label{appendix computational details}
\textbf{Extract input vector $x$ with token average}.
In our method, we extract LoRA increment $\mathbf{d}Wx$ as the functional direction, where $x$ is the input \textit{vector} of the module with the parameter matrix $W$. Note that in a transformer model, the input, say $X$, to $W$ is several vectors, that is:
\begin{equation}
    X = (x_1, x_2,...,x_n)
\end{equation}
where $N$ is the number of input tokens, $x_n$ is the input vector at the place of the $n$-th token. Common methods to represent inputs $x_1, x_2,...,x_n$ with a single vector $x$ include computing their average or taking the last vector. For stability of computation, we employ the average method, that is:
\begin{equation}
    x = \frac{1}{N}\sum_{n=1}^N x_n
\end{equation}

\textbf{Standarization of LoRA increment $h$ for PCA}.
We employ the LoRA increment $h$(computed with equation~\eqref{onlinecl-data}) as the functional direction in our method. 
As there is no scale difference in gradients, we omit normalization, following~\citep{pcasurvey}. Note that we are concerned with only the directions of $h$, 
so we also conduct no centralization for $h$ at the beginning. Note that in this case, the first few principal component represents the weighted historical average~\eqref{ccipca:1}, and the residual cut in equation~\eqref{ccipca-residual} will deduct the average and thus reach certain centralization. Also, the effect of other normalization methods designed for LoRA increments or gradients deserves further investigation.

\section{Additional Related works}\label{appendix related works}

The following methods have been developed for LLM continual learning. They can
be categorized into the following types: Rehearsal-based
%~\citep{rehearsal1, rehearsal2, rehearsal3}
, Architecture-based
%\citep{expert1, expertavg, llama-pro}
, Prompt-based,
%~\citep{l2p,lfpt5,progressive-prompt}
and Regularization-based approaches.
%~\citep{ogd, olora, lwf, ewc, synapsis}
A brief summary is in Table~\ref{method-comparisons}.

\textbf{Rehearsal-based approach}
\citep{rehearsal1,rehearsal2,rehearsal3} try to remind the model of historical tasks and thus avoid forgetting. However, there are growing restoration costs as tasks accumulate, and privacy issues in gaining historical training data.

\textbf{Architecture-based approach}
\citep{expert1,expert2,expertavg} train multiple expert models for each task. However, when it comes to unseen tasks, there is no proper expert to use, which destroys the generalization ability of models. 

\textbf{Prompt-based approach}
L2P\citep{l2p}, LFPT5\citep{lfpt5}, and Progressive Prompts\citep{progressive-prompt} add prompts during the inference of the model. This approach is lightweight, but when the fine-tuning information gets large, the prompt will not be able to cover it.

\textbf{Regularization-based Approach}
EWC\citep{ewc, synapsis}, LwF\citep{lwf}, OGD\citep{ogd}, and O-LoRA\citep{olora} limit the update of model parameters to preserve the historical ability of the model. Their edge is that no historical data or extra architecture is required. We lay emphasis on the orthogonal methods, including OGD and O-LoRA, as we employ orthogonal cuts to avoid changing historical parameter settings and preserve historical functions.

\textbf{Orthogonal methods}
The key point of orthogonal methods is to avoid wrecking the parameter subspace of historical tasks when fine-tuning on the latest task, and the method is to make the parameter space of new tasks orthogonal to the historical ones.
Representative methods, including Orthogonal Gradient Descent (OGD)\citep{ogd} and Orthogonal Subspace Learning(O-LoRA)\citep{olora}, have been proven effective in preventing catastrophic forgetting. 
OGD forces the gradient descent to be orthogonal to the gradient directions of historical tasks. That is:
\begin{equation}
    G_T \perp G_t  \quad t=1, 2,..., T-1
\end{equation}
where $G_t = \nabla_{\theta} L_t$ is the gradient direction of the $t$ th task. O-LoRA tries to make the LoRA $B$ matrix in equation~\eqref{lora} orthogonal to that of historical LoRA modules. That is:
\begin{equation}
    \beta_T^i \perp \beta_t^j \quad t=1,2,...T-1 \quad i,j=1,2,...,r
\end{equation}
where $\beta_t^i$ is the $i$ th colomun vector of $B_{m \times r}$ matrix fine-tuned in the $t$ th task, that is $B = (\beta_1, \beta_2,..., \beta_r)$.

\begin{table}[t]
\centering
\caption{The accuracy on the MMLU benchmark of LLaMA-7B before and after continual learning (CL) on the CL benchmark. The results are the average of task orders 1-3. Note that with MMLU being a four-classification problem, a 25\% accuracy equates to random guessing.}
\begin{tabular}{c | c}
\toprule
 & \textbf{MMLU Accuracy} \\
\midrule
Original model            & 32.3        \\
Alpaca LoRA fine-tuned model     & 36.0     \\
\midrule
Seq LoRA CL after Alpaca LoRA      & 26.2      \\
O-LoRA CL after Alpaca LoRA   & 30.1     \\
DOC CL after Alpaca LoRA   & 29.4     \\
\midrule
O-LoRA throughout Alpaca and CL      & 32.1 \\ 
\textbf{DOC throughout Alpaca and CL}      & \textbf{34.6}     \\
\bottomrule
\end{tabular}

\label{tab-mmlu}
\end{table}

\begin{table}[t]
\centering
\caption{The comparison of continual learning methods. Specifically, \textbf{RF} indicates whether the method is rehearsal-free. \textbf{TIF} indicates whether the task ID is free during inference. Compared to regularization-based methods, other methods have extra settings or computational overheads.}
\resizebox{\textwidth}{!}{
\begin{tabular}{l | l cc l}
\toprule
                
               & & \textbf{\ RF\ } &  \textbf{\ TIF\ } & \textbf{\ Inference costs\ } \\

\midrule

\multirow{2}{*}{Rehearsal-based} &MBPA++ \citep{rehearsal1}         &                                                        & $\checkmark$      

\\
&IDBR  \citep{rehearsal3}          &                                                        & $\checkmark$                                              \\
\midrule

\multirow{4}{*}{Architecture-based}
&EIP~\citep{expert4} & $\checkmark$ & $\checkmark$ &
\multirow{4}{*}{Expert selection}
\\

&SLM~\citep{expertavg} & $\checkmark$ & $\checkmark$ \\

&Expert LMs~\citep{expert1} &  & $\checkmark$ \\

&MoCL~\citep{expert2} & $\checkmark$ & $\checkmark$\\

%&LLaMA Pro~\citep{llama-pro} & $\checkmark$ & $\checkmark$ \\

\midrule
\multirow{3}{*}{Prompt-based}
&L2P \citep{l2p}            & $\checkmark$                                                 & $\checkmark$    &                                          
\multirow{3}{*}{Additional prompts}
\\

&LFPT5 \citep{lfpt5}          &                         & $\checkmark$                                                                           \\

&ProgPrompt  \citep{progressive-prompt}            & $\checkmark$                                  &                                             \\  

\midrule
\multirow{4}{*}{Regularizatoin-based}
&EWC \citep{ewc}            & $\checkmark$                                                      & $\checkmark$                                               \\

&LwF \citep{lwf}            & $\checkmark$                                                      &                                                 \\
&OGD \citep{ogd}            & $\checkmark$                                                      & $\checkmark$                                        \\

 & O-LoRA~\citep{olora} & $\checkmark$                                          & $\checkmark$                                           \\ 
 & \textbf{DOC}(ours) & $\checkmark$                                          & $\checkmark$                                           \\ 
\bottomrule

%\specialrule{2pt}{0pt}{\belowrulesep}
\end{tabular}
}

\label{method-comparisons}
\end{table}

\section{Limitations}\label{limitations}
While the proposed method has an outstanding performance in empirical evaluations, we discuss its potential limitations as follows. 

\begin{figure}[t]
    \centering
    \begin{tabular}{c}
    \includegraphics[width=0.8\textwidth]{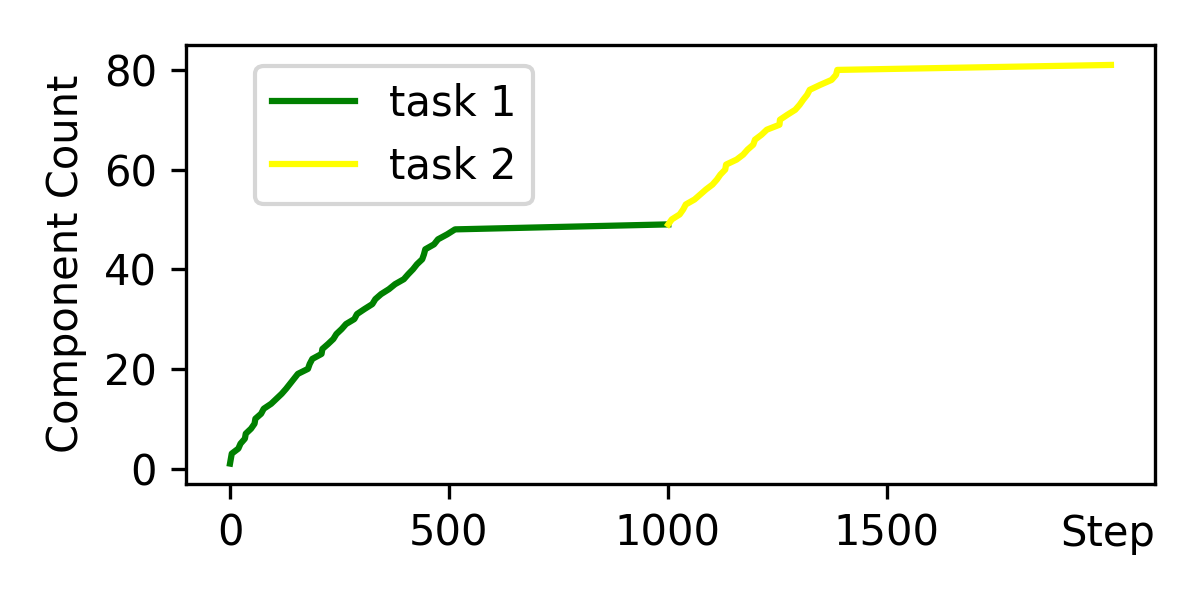} \\
    \end{tabular}
    \vspace{-10pt}
    \caption{The expansion of principal components in the first 2 tasks. Note that we do not limit the maximum principal component number in this experiment, \textit{i.e.}, $K_\text{max} = +\infty$. As the number of principal components increases, it reaches a point where no extra component is required, indicating that the current components are adequate to cover a large enough part of the LoRA increment.
    %as presented in Algorithm~\ref{ccipca-algorithm}. 
    }
    \label{fig-PCnum}
\end{figure}

\textbf{Scalability}.
 In more complex scenarios with a large number of tasks, such as hundreds of tasks, the principal component pool expands as we add new components during the fine-tuning process, imposing a growing load for computation. The empirical scale of the expansion is approximately 40 components for each task, as shown in Figure~\ref{fig-PCnum}. The size of these components is approximately 15MB and is acceptable in the settings of our experiments. 
 However, in the case of hundreds of tasks, the performance and applicability of our method require further investigation. 

\textbf{Task identification}.
 Although our method requires no task identification during inference, it is still required during the continual fine-tuning process. Exploring methods for task-agnostic training would be a valuable future direction. This is further discussed in Future Directions.

\textbf{Generalization Ability}.
 As our method is targeted at preserving historical functions, it has no recognition of unseen tasks. Its generalization ability deserves further investigation. We propose the following empirical demonstration of the impact of our method on the generalization ability of the model.
 Following O-LoRA~\citep{olora}, we start with a fine-tuned LLaMA-7B language model on the Alpaca~\citep{alpaca} dataset. After conducting continual learning on the CL benchmark~\citep{clbenchmark}, we test the model on the MMLU benchmark~\citep{mmlu, ethics}, composed of unseen tasks. The results are shown in Table~\ref{tab-mmlu}.   
 Compared to the original model, SeqLoRA and our method (DOC) suffer from forgetting (accuracy respectively drops from 36.0 to 26.2 and 29.4). This is because of the lack of information about unseen tasks during continual learning. In the experimental settings, the issues above limit the practicality of DOC.

 Note that we fine-tune the model on Alpaca at the beginning, so that continual learning also triggers the forgetting of Alpaca. What if we mitigate this forgetting with CL methods? We further investigate the effect of continual fine-tuning the model on the Alpaca and CL benchmark with CL methods applied \textbf{throughout from start to end}, which makes the Alpaca \textbf{visible} to the methods. Note that MMLU is still unseen during the continual fine-tuning process in this setting. The results show the enhanced performance (an accuracy of 34.6 for DOC and 32.1 for O-LoRA, compared to 29.4 and 30.1 in the former experiment where Alpaca is invisible to the methods). An explanation is that the methods avoid forgetting Alpaca, which is a general dataset that assists in the initialization of crucial functional directions of the model, thus aiding in the preservation of crucial functions for unseen tasks. The results also indicate that the generalization of our method, which preserves the ability on unseen tasks with a general dataset, is better compared to O-LoRA. It inspires the practical deployment of DOC to initialize on a general dataset beforehand.

\section{Future directoins}\label{future directions}

\textbf{The interpretability of principal components}.
We employ PCA in our method for functional direction tracking. Another edge of PCA is that the components extracted are statistically independent of each other; thus, each component represents an individual unit, as proposed by~\citep{quantom}. The individuality of these components provides chances for model deconstruction and better interpretability, and it is possible to find the exact meaning of each component, \textit{e.g.}, semantic function, logic function, certain knowledge, \textit{etc.},  through empirical methods. It is a promising direction for interpretable learning based on model deconstruction.

\textbf{Automated task ID recognition}
As mentioned before, exploring methods for task-agnostic training would be valuable. It deserves further investigation into the characteristics of the principal components extracted from a specific task, which assists in the distinction of different tasks.

\end{document}